\pdfoutput=1
\documentclass[11pt]{article}
\usepackage[dvipsnames]{xcolor}
\usepackage{graphicx}
\usepackage[]{acl}

\usepackage{times}
\usepackage{latexsym}
\usepackage{tikz}
\usepackage{pgfplots}
\usepackage[T1]{fontenc}
\usepackage[utf8]{inputenc}

\usepackage{microtype}

\usepackage{comment}
\usepackage{amsfonts}
\usepackage{amsmath}
\usepackage{siunitx}
\usepackage{booktabs}
\usepackage{tabularx}
\usepackage{multirow}
\usepackage{booktabs}
\usepackage[export, demo]{adjustbox}  
\usepackage{multicol, makecell}
\usepackage{bbm}
\usepackage{tablefootnote}
\usepackage{romannum}
\usepackage{subcaption}
\usepackage{tikz}

\newcolumntype{C}{>{\centering\arraybackslash}X}

\definecolor{light-gray}{gray}{0.95}

\usepackage{pifont}% http://ctan.org/pkg/pifont

\usepackage{listings}
\usepackage{xcolor}

\lstdefinestyle{mystyle}{
    basicstyle=\ttfamily\small,
    frame=single,
    captionpos=b,
    breaklines=true,
    breakindent=0pt,          % No indentation on broken lines
    breakautoindent=false,    % Disable auto-indentation
    backgroundcolor=\color{gray!2},
}

\definecolor{olive}{RGB}{0,153,51}

\title{Robust Adaptation of Large Multimodal Models for Retrieval Augmented Hateful Meme Detection}

\author{Jingbiao Mei, Jinghong Chen, Guangyu Yang, Weizhe Lin, Bill Byrne \\
Department of Engineering\\
University of Cambridge\\
Cambridge, United Kingdom, CB2 1PZ \\
  \texttt{\{jm2245, jc2124, gy266, wl356, wjb31\}@cam.ac.uk} \\}
  
\begin{document}
\pagenumbering{arabic}
\maketitle

\begin{abstract}
Hateful memes have become a significant concern on the Internet, necessitating robust automated detection systems. While Large Multimodal Models (LMMs) have shown promise in hateful meme detection, they face notable challenges like sub-optimal performance and limited out-of-domain generalization capabilities. Recent studies further reveal the limitations of both supervised fine-tuning (SFT) and in-context learning when applied to LMMs in this setting. To address these issues, we propose a robust adaptation framework for hateful meme detection that enhances in-domain accuracy and cross-domain generalization while preserving the general vision-language capabilities of LMMs. Analysis reveals that our approach achieves improved robustness under adversarial attacks compared to SFT models. Experiments on six meme classification datasets show that our approach achieves state-of-the-art performance, outperforming larger agentic systems.
Moreover, our method generates higher-quality rationales for explaining hateful content compared to standard SFT, enhancing model interpretability. Code available at \href{https://github.com/JingbiaoMei/RGCL}{https://github.com/JingbiaoMei/RGCL}
\end{abstract}

 \textcolor{red}{This paper contains content for demonstration purposes that may be disturbing for some readers.}

\section{Introduction}

The rise of social media has led to a surge in hateful content, notably in the form of memes. Manual detection is infeasible due to the vast amount of content and psychological risks for human moderators. Consequently, hateful meme detection systems have attracted considerable research interest \cite{KielaFBHMC2020, LiuFigMemes2022, PrakashTotalDefMeme2023, Shah2024memeclip_pridemm}. 

Large Multimodal Models (LMMs) have emerged as a promising solution for this complex task \cite{Hee2024RecentAdvanceHateSpeech, Lin2025GoatBench}. Their strong capabilities across a range of general vision-language tasks provide a solid foundation for understanding the intricate interplay between text and image in memes \cite{Zhu_MiniGPT4_2023, LiuLLAVA2023}. Furthermore, the generative nature of LMMs offers interpretability, allowing models to provide rationales for their detection decisions. Ideally, LMMs also bring improved generalizability, enabling them to adapt to the rapidly evolving landscape of online memes, making them well-suited for deployment in real-world content moderation systems. 
Despite this potential, current LMMs face the following challenges when applied to hateful meme detection.
\begin{enumerate}
    \item \textbf{Sub-optimal performance.} %Standard supervised fine-tuning (SFT) and in-context learning with decoder‐only LMMs perform worse than CLIP‐based methods \cite{clip2021} on meme classification datasets \cite{RGCL2024Mei, Hee2024BridgeModality}. 
    %Standard supervised fine-tuning (SFT) results in sub-optimal performance for LMMs, as it struggles to learn the complex interplay of visual and textual cues inherent in hateful memes \cite{RGCL2024Mei}. As further demonstrated in our analysis in Appendix~\ref{appendix:rationale}, SFT models also produce lower-quality rationales when explaining hateful content, highlighting their limitations in modeling the nuanced nature of hateful memes.
    LMMs struggle to learn the interplay of visual and textual cues inherent in hateful memes through standard supervised fine-tuning (SFT), as reported by \citet{RGCL2024Mei}. We also report that SFT LMMs produce lower-quality rationales when explaining hateful content, possibly caused by overfitting and the scarcity of the training data.
    
    \item \textbf{Limited out-of-domain generalization.} Memes constantly evolve with social trends and events, posing a generalization challenge \cite{Cao_2024_ModHate, RGCL2024Mei}. While in-context learning with retrieved examples from a dynamic meme database is a potential approach to generalize to unseen data for LMMs, \citet{Huang_LowResourceLMMAgentHatefulMeme_2024} found that this approach remains ineffective, highlighting the need for more effective methods to use few-shot meme examples.
    \item \textbf{Degradation of general vision-language abilities can arise from fine-tuning for meme classification.} We observe that applying SFT for meme classification leads to overfitting, which degrades performance on general multimodal benchmarks like MMMU \cite{Yue2023MMMU}. This undermines the rationale for choosing LMMs over single-purpose specialized models such as CLIP.
    
\end{enumerate}

%RA-HMD enables Qwen2-VL-7B \cite{Wang_2024_Qwen2VL} to achieve state-of-the-art performance on six meme classification datasets under supervised settings. 
%Furthermore, the retrieval-augmented KNN classifier within RA-HMD leverages few-shot examples more effectively than conventional in-context learning frameworks for robust out-of-domain generalization. As shown in Section~\ref{sec:results_lowres}, RA-HMD enhanced Qwen2-VL-7B surpasses models such as GPT-4o \cite{OpenAI2024gpt4o} and LOREHM \cite{Huang_LowResourceLMMAgentHatefulMeme_2024} with LLaVA-34B \cite{Liu_2023_LLAVA1.5} under out-of-domain settings. 

%As noted, LMMs are capable of generating rationales for their decisions. RA-HMD produces higher-quality rationales for hateful meme classification compared to SFT, as shown in Figure~\ref{fig:pairwise_compare}.
%Furthermore, results in Section~\ref{sec:general} show that RA-HMD more effectively preserves the general vision-language capabilities of LMMs than SFT.

\begin{figure}[h]
\centering
\begin{tikzpicture}[scale=1.1]

    % Total width of bar = 6.5cm
    \def\totalwidth{6.5}
    \def\barheight{0.6}

    % Define relative widths (16%, 14%, 70%)
    \def\wA{0.9}  % 16% of 6.5
    \def\wB{1.6}  % 14% of 6.5
    \def\wC{4.0}  % 70% of 6.5

    % Draw sections
    \fill[red!70!black] (0,0) rectangle (\wA,\barheight);
    \fill[orange!90!black] (\wA,0) rectangle ({\wA+\wB},\barheight);
    \fill[green!60!black] ({\wA+\wB},0) rectangle (\totalwidth,\barheight);

    % Add text labels inside the bars
    \node at ({0.5*\wA},0.3) {\textbf{\textcolor{white}{14\%}}};
    \node at ({\wA + 0.5*\wB},0.3) {\textbf{\textcolor{white}{25\%}}};
    \node at ({\wA + \wB + 0.5*\wC},0.3) {\textbf{\textcolor{white}{61\%}}};

    % Legend below
    \begin{scope}[yshift=-0.6cm]
        \draw[fill=red!70!black] (0,0) rectangle ++(0.3,0.3);
        \node[right=2pt] at (0.4,0.2) {\small SFT};

        \draw[fill=orange!90!black] (2,0) rectangle ++(0.3,0.3);
        \node[right=2pt] at (2.4,0.2) {\small Tie};

        \draw[fill=green!60!black] (4,0) rectangle ++(0.3,0.3);
        \node[right=2pt] at (4.4,0.2) {\small RA-HMD};
    \end{scope}

\end{tikzpicture}
\caption{Comparison of rationales generated by SFT and RA-HMD Qwen2-VL-7B models on the HatefulMemes dataset. The bar chart shows the winning rate of rationale quality based on pairwise comparisons between the two models. A more detailed analysis is provided in Appendix~\ref{appendix:rationale}. }
\label{fig:pairwise_compare}
\end{figure}

%In summary, our contributions are:
To address these challenges, we propose \textbf{RA-HMD} (\textbf{R}etrieval-\textbf{A}ugmented \textbf{H}ateful \textbf{M}eme \textbf{D}etection), a framework that incorporates architectural enhancements and a two-stage fine-tuning strategy to adapt LMMs for hateful meme detection without degradation in general vision-language ability. 

We address the \textbf{three challenges} of applying LMMs to hateful meme classification through the following contributions:
\begin{enumerate}

\item  We propose RA-HMD, a fine-tuning framework for adapting LMMs for hateful meme classification, achieving new state-of-the-art results on six widely used meme classification datasets. In addition, RA-HMD generates higher quality rationales compared to SFT models,  thereby enhancing the interpretability of LMM predictions, as shown in Figure~\ref{fig:pairwise_compare}.
\item RA-HMD demonstrates more robust out-of-domain generalization compared to SFT models. Notably, when RA-HMD is combined with a retrieval-augmented KNN classifier, it demonstrates state-of-the-art performance for out-of-domain meme classification. Moreover, this setup enhances robustness against adversarial attacks and leverages few-shot meme examples more effectively than in-context learning, thereby addressing the challenge of adapting to rapidly evolving memes without the need for retraining.
\item %RA-HMD preserves the general vision–language capabilities of LMMs, making it an effective and robust post-training enhancement for general-purpose multimodal models.
%expanding their abilities to do meme classification while retaining their other abilities. without loss in other skills.
RA-HMD expands LMMs ability to perform hateful meme classification and explaining hateful memes without compromising performance on other vision–language tasks, as shown by results in Section~\ref{sec:general}.

%\item Our system generates higher quality rationales compared to SFT models, enhancing the interpretability of LMM predictions in the context of hateful meme detection, as shown in Figure~\ref{fig:pairwise_compare}.
\end{enumerate}

\section{Related Work}
%\subsection{Hateful Memes Detection}
%The Hateful Memes Challenge competition \cite{KielaFBHMC2020} released a benchmark dataset for hateful meme detection. 
%The best baseline model from the challenge is the Visual BERT, achieving an AUROC of 75.4. The prize-winning solution \cite{RonHMC1st2020, VilioHMC2nd2020, RizaHMC3rd2020} of the challenge managed to push the AUROC to 84.5 with additional extracted features and ensembling of models.
\subsection{Hateful Meme Detection}
Most existing approaches to hateful meme detection rely on supervised learning, with the majority of research leveraging CLIP \cite{clip2021}. Numerous studies have fine-tuned models based on CLIP using different modality fusion mechanisms~\cite{PramanickMomenta2021, KumarHateClip2022, Shah2024memeclip_pridemm}. Other works incorporate caption models into the CLIP-based feature fusion network to further enhance performance~\cite{Burbi_2023_Issues, Cao_2023_ProCap, Ji2024CapAlign}. Additionally, contrastive learning techniques have been explored to address confounding factors in meme classification~\cite{LippeHMFramework2020, RGCL2024Mei}. 

With the emergence of LMMs, recent research has shifted toward using LMMs as generalist models, in contrast to the specialist nature of CLIP based models \cite{Laurençon_2023OBELICS, Hu_2024_VPD}. Moreover, decoder-based LMMs offer an additional advantage: they can generate textual rationales to explain why a meme may be hateful \cite{Lin_2024_ExplainHM, Hee2024RecentAdvanceHateSpeech}.

While LMMs such as Flamingo~\cite{Flamingo22} have shown promise in hateful meme detection via SFT, fine-tuning strategies for LMMs remain underexplored. In fact, \citet{RGCL2024Mei} demonstrated that fine-tuned CLIP models can outperform much larger LMMs, highlighting the need for specialized methods. In this work, we address this gap by proposing LMM architecture refinement alongside a novel fine-tuning approach for LMMs that enhances their performance on hateful meme detection while preserving their general vision–language capabilities.

\subsection{Low resource hateful meme detection} 
Low-resource hateful meme detection is critical for real-world applications that demand out-of-domain generalization. In this setting, an initially trained model is deployed to a new domain without gradient updates, relying only on demonstration examples for inference \cite{Huang_LowResourceLMMAgentHatefulMeme_2024}.  \citet{Hee2024BridgeModality} utilized retrieved few-shot examples to help LMMs generalize to unseen memes. \citet{Hu_2024_VPD} and \citet{Huang_LowResourceLMMAgentHatefulMeme_2024} explored agent-based LMM systems with few-shot learning for out-of-domain settings. However, \citet{Huang_LowResourceLMMAgentHatefulMeme_2024} observed that in-context learning is less effective for meme classification compared to other tasks, highlighting the need for more effective strategies to use demonstration examples.
\begin{figure*}[h!]
    \centering
    \includegraphics[width=\textwidth]{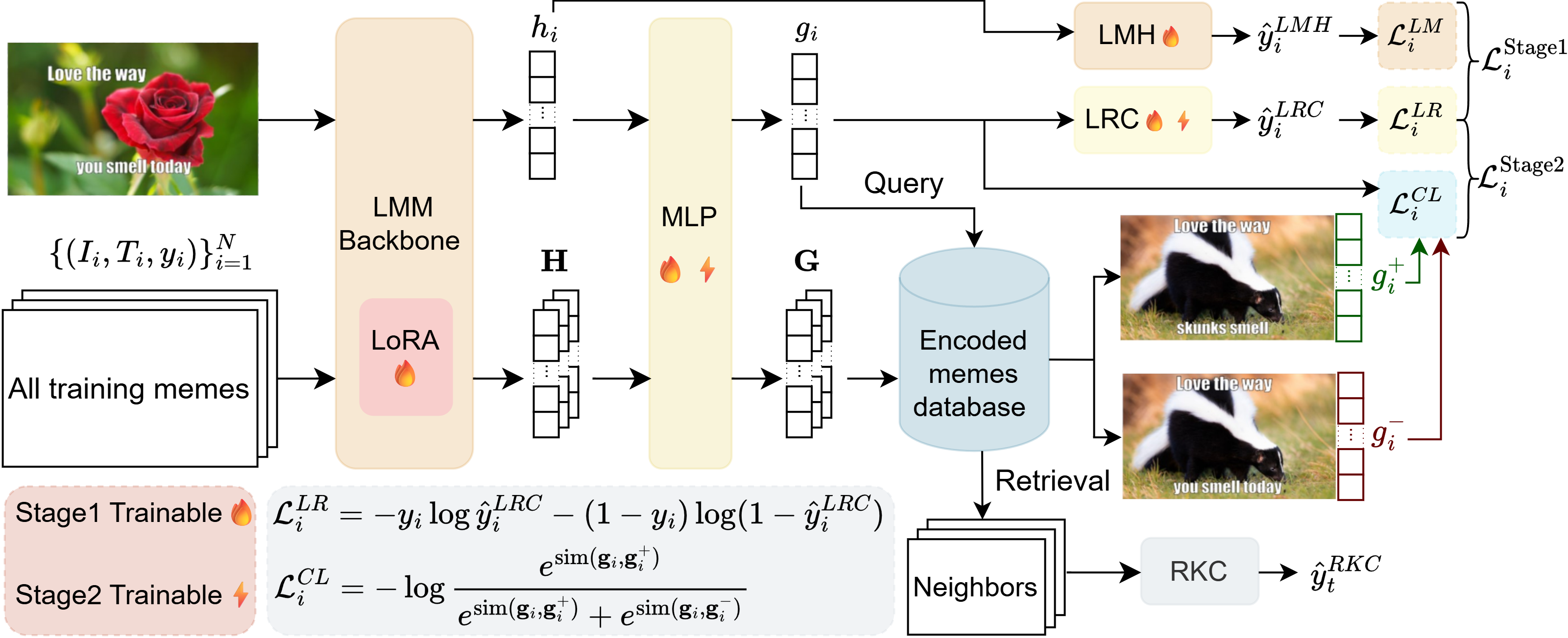}
    \caption{Architecture of RA-HMD.
    We decompose the LMM into two components: the LMM Backbone and the LM Head (LMH). For each training example $i$, the last hidden state $\mathbf{h}_i$ is fed to the LMH to obtain the LM loss $\mathcal{L}_i^{LM}$. $\mathbf{h}_i$ is also fed to a trainable multilayer perceptron (MLP) to generate an embedding $\mathbf{g}_i$ for use as a retrieval query and as a feature for the Logistic Regression Classifier (LRC) to compute the cross entropy loss $\mathcal{L}_i^{LR}$. During training, contrastive learning examples are retrieved from the encoded meme database $\mathbf{G}$ for computing the contrastive loss $\mathcal{L}_i^{CL}$. At inference, the same process retrieves the $K$ nearest neighbors for Retrieval-augmented KNN Classification (RKC),  which predicts the label $\hat{y}_t^{RKC}$ for an inference example $t$.}
    \label{fig:system}
\end{figure*}
%In contrast, we use a retrieval-augmented approach for classifying unseen memes and find that it makes more effective use of demonstration examples than conventional in-context learning.
In this work, we show that RA-HMD improves the in-context learning capabilities of LMMs. Furthermore, when combined with a retrieval-augmented KNN classifier, RA-HMD enables more effective use of demonstration examples than conventional in-context learning.
\section{RA-HMD Methodology}
\subsection{Preliminaries}
\label{sec:preliminaries}
\textbf{Problem Statement } Hateful memes datasets are defined as
$\{ (I_i,T_i,y_i)\}_{i=1}^N$, %$\{ (I_i,T_i)\}_{i=1}^N$
where $I_i \in \mathbb{R}^{C\times H \times W }$ is the image portion of the meme in pixels; $T_i$ is the caption overlaid on the meme; $y_i\in\{0,1\}$ is the label, where 0 stands for benign, 1 for hateful.
\\
\textbf{Large Multimodal Models}
Some prior work has approached hateful meme detection via text generation with LMMs, where the LMM takes a meme $(I_i, T_i)$ as an input to predict a single token label $\hat{y}_i^{LMH} \in \{\textrm{``benign'', ``hateful''}\}$~\cite{Lin_2024_ExplainHM}.
We refer to the final linear layer of the LMM as the LM Head (\textbf{LMH}), which maps hidden representations to a probability distribution over the vocabulary via a softmax function. For meme classification, the LMH decodes the hidden state of the last token and generates the output label.
This contrasts with approaches based on CLIP, which train Logistic Regression Classifiers \textbf{(LRC)} on encoder CLS tokens \cite{KumarHateClip2022}.

%\textbf{Supervised Fine-Tuning (SFT).} SFT adapts LMMs to downstream tasks by optimizing the language modeling objective $\mathcal{L}_i^{LM}$, the negative log-likelihood of the next token prediction.  Our framework, RA-HMD, extends this paradigm with a two-stage fine-tuning strategy to enhance both classification and retrieval capabilities.

%\textbf{Low resource Hateful Memes Detection}
%Following previous works \cite{Huang_LowResourceLMMAgentHatefulMeme_2024}, we define a 
\subsection{RA-HMD Framework}
\textbf{Architecture enhancement} 
Leveraging representations from large multimodal models (LMMs) for hateful meme classification is non-trivial, particularly when attempting to use LMM embeddings for classification while preserving the model’s original language generation capabilities. 

Prior work has explored various strategies for adapting these representations to classification and retrieval tasks. In our study, we similarly experimented with multiple adaptation methods. Appendix~\ref{appendix:insights_representation} provides a comprehensive summary of these efforts, including failure cases of previous approaches and key insights that ultimately guided the design of RA-HMD. 

A central takeaway is that earlier adaptation methods enabled LMMs to perform retrieval but failed to preserve their ability to generate text simultaneously. In contrast, our proposed architecture, combined with a two-stage training procedure, successfully addresses this limitation.

As illustrated in Figure~\ref{fig:system}, RA-HMD integrates an LMM with two additional trainable components: a Multilayer Perceptron (MLP) that projects the LMM final hidden state $\mathbf{h}_i$ into an embedding $\mathbf{g}_i$ for use in classification and retrieval; and an LRC
operating on $\mathbf{g}_i$ . Figure~\ref{fig:system} shows how the architecture supports multiple fine-tuning and inference modes.

\noindent\textbf{Retrieval} During stage-2 training, FAISS-based~\cite{johnson_Faiss2019billion} nearest neighbor search retrieves contrastive learning examples from the encoded meme database $\mathbf{G}$. At inference, FAISS is used to retrieve neighbors for the Retrieval-augmented KNN Classifier (\textbf{RKC}).

\noindent\textbf{Inference modes} 
Figure~\ref{fig:system} shows three different classifiers: LMH, LRC, and RKC.
%1. The LM head provides baseline inference results for pre-trained and SFT LMMs, following prior work (Section~\ref{sec:preliminaries}). 
%2. The \howard{LRC} is our default classifier in supervised settings. 3. RKC is our default classifier in out-of-domain settings. 
For pre-trained and SFT LMMs, we generate classification decisions using the LMH as described in Section~\ref{sec:preliminaries}. For RA-HMD models, we obtain meme classification decisions via the LRC, unless otherwise specified. 
Section~\ref{sec:abl_infer_mode} presents a detailed comparison of the three inference modes.
\subsection{Stage 1: Logistic Regression Augmented Supervised Fine-tuning}
In stage 1, the LMM is fine-tuned via Low-Rank Adaptation~\cite{Hu_LORA_2021},  which applies trainable low-rank matrices to the model while freezing its original weights. The MLP and LRC are updated simultaneously.
We optimize the joint loss for each training example $i$:
\begin{equation}
\mathcal{L}_{i}^{\textrm{Stage1}}=
\mathcal{L}_i^{LM} +\mathcal{L}_i^{LR},
\label{eq:stage1}
\end{equation}
%$\howard{where \mathcal{L}_i^{LM}$ is the language modeling objective used in SFT, and the $\mathcal{L}_i^{LR}$ is the binary cross entropy loss applied to the LRC prediction $\hat{y}_i^{LRC}$:}
where $\mathcal{L}_i^{LM}$ is the language modeling objective used in SFT. In the context of meme classification, the model is trained to predict a single target token $s(y_i)$:
\begin{equation}
    s(y_i)     =      \begin{cases}
        \textrm{``benign''} &\text{if } y_i=0\\
        \textrm{``hateful''}  &\text{if } y_i= 1

    \end{cases}.
\end{equation}
$\mathcal{L}_i^{LM}$ is computed as the negative log-likelihood of generating the correct target token, conditioned on the input image and text:
\begin{equation}
    \mathcal{L}_i^{LM} = -\log p(\hat{y}_i^{LMH} =s(y_i)  \mid I_i, T_i)   
\end{equation}
The $\mathcal{L}_i^{LR}$ is the binary cross-entropy loss applied to the LRC prediction $\hat{y}_i^{LRC}$:
\begin{equation}
\mathcal{L}_i^{LR} = - y_i\log \hat{y}_i^{LRC} - (1-y_i)\log(1-\hat{y}_i^{LRC})
\label{eq:CE}
\end{equation}
%Jointly optimizing the \textit{LM} loss with the \textit{CE} loss allows the LMM to rapidly adapt to the hateful meme detection task while preserving its generative capabilities.
Jointly optimizing the language modeling loss $\mathcal{L}_i^{LM}$ with the cross-entropy loss $\mathcal{L}_i^{LR}$ allows the LMM to rapidly adapt to the hateful meme detection task.

\subsection{Stage 2: LMM Contrastive Fine-tuning}
In stage 2, the LMM is frozen; only the MLP and LRC are fine-tuned to refine retrieval-aligned representations.
%We further improve the system’s retrieval and classification performance by fine-tuning the MLP and the \howard{\textbf{LRC}}  while keeping the LMM fixed. 
Stage 2 jointly optimizes:
\begin{equation}
    \mathcal{L}_i^{\textrm{Stage2}} = \mathcal{L}_i^{CL} + \mathcal{L}_i^{LR},
\label{eq:stage2}
\end{equation}
where $\mathcal{L}_i^{LR}$ is defined in Eq.~\ref{eq:CE}, and \(\mathcal{L}_i^{CL}\) is the Contrastive Learning Loss.

To compute \(\mathcal{L}_i^{CL}\), we retrieve pseudo-gold positive examples similar to RGCL \cite{RGCL2024Mei} and hard negative examples \cite{Schroff_FaceNet_2015} from the training set. Specifically, for a given sample \(i\) with embedding \(\mathbf{g}_i\), we use FAISS \cite{johnson_Faiss2019billion} to perform the nearest neighbor search between \(\mathbf{g}_i\) and every other target embedding \(\mathbf{g}_j \in \mathbf{G}\) from the training set. %The encoded meme database $\mathbf{G}$ is updated every 100 steps during fine-tuning. 

Pseudo-gold positive examples are same-label examples that have high similarity scores with \(\mathbf{g}_i\), while hard negative examples are opposite-label examples that have high similarity scores. We denote the embedding of the pseudo-gold positive example and hard negative example as \(\mathbf{g}_i^{+}\) and \(\mathbf{g}_i^{-}\), respectively.
\(\mathcal{L}_i^{CL}\) is then computed as:
\begin{align}
\nonumber \mathcal{L}_i^{CL}&= L(\mathbf{g}_i,  \mathbf{g}_i^{+}, \mathbf{g}_{i}^{-})\\
        &= - \log \frac{ e^{\textrm{sim}(\mathbf{g}_i,\mathbf{g}_{i}^{+})}}{ e^{\textrm{sim}(\mathbf{g}_i,\mathbf{g}_{i}^{+})} + e^{\textrm{sim}(\mathbf{g}_i,\mathbf{g}_{i}^{-})}},
\end{align}
where \(\textrm{sim}(\cdot,\cdot)\) denotes the cosine similarity function. Stage 2 fine-tuning explicitly aligns the representations of semantically similar meme pairs, thereby improving the generalization of LMMs to distribution shifts in unseen datasets.

\subsection{Retrieval Augmented KNN Classification}
In addition to the LMH and LRC, RKC is used specifically for out-of-domain meme classification. 
For a test meme $t$, we retrieve $K$ similar memes within the embedding space from the meme database $\mathbf{G}$.
We perform similarity-weighted majority voting to obtain the prediction:
\begin{equation}
    \hat{y}_t^{RKC} = \sigma(\sum_{k=1}^K\overline{y}_k \cdot \text{sim}(g_k, g_t)),
\end{equation}
where $\sigma(\cdot)$ is the sigmoid function and 
\begin{equation}
    \overline{y}_k:=
    \begin{cases}
        1  &\text{if } y_k= 1\\
        -1 &\text{if } y_k=0
    \end{cases}.
    \label{eq:indicator_function}
\end{equation}
Additionally, to enable RKC on pretrained or SFT LMMs that do not incorporate an MLP, we use the last hidden state \(\mathbf{h}_i\) for the nearest neighbor search. The results are provided in Appendix~\ref{appendix:rkc_icl_ft}.
\section{Experiments}
We evaluate on six meme classification datasets: \textbf{HatefulMemes}~\cite{KielaFBHMC2020}, \textbf{HarMeme}~\cite{pramanickCovidMeme2021}, \textbf{MAMI}~\cite{Fersini_MAMI_2022}, \textbf{Harm-P}~\cite{PramanickMomenta2021}, \textbf{MultiOFF}~\cite{suryawanshi-etal-2020-MultiOFF} and \textbf{PrideMM}~\cite{Shah2024memeclip_pridemm}. %These datasets encompass varying definitions of harmful content across different sociopolitical contexts.
A detailed description and statistics are in Appendix~\ref{appendix:data_stats}. %For HatefulMemes, HarMeme, and MAMI, we report the Area Under the Receiver Operating Characteristic Curve (AUC) and Accuracy (Acc) in line with previous studies \cite{KumarHateClip2022, Cao_2023_ProCap, RGCL2024Mei, Cao_2024_ModHate}. For Harm-P, MultiOFF, and PrideMM, we report Accuracy and F1 score consistent with the literature \cite{PramanickMomenta2021, RGCL2024Mei, Shah2024memeclip_pridemm, Lin_2024_ExplainHM}.
Implementation details are described in Appendix~\ref{appendix:exp_setup}.

\begin{table*}[h]
\sisetup{table-format=2.1, table-auto-round=true, table-number-alignment=center, detect-weight=true, detect-family=true,mode=text}
\small
\centering
\setlength{\tabcolsep}{4pt}{
\resizebox{1.0\linewidth}{!}{%
\begin{tabular}{ll|SS|SS|SS|SS|SS|SS}
\toprule 
\multicolumn{2}{c|}{} &
 \multicolumn{2}{c|}{\textbf{HatefulMemes}} & \multicolumn{2}{c|}{\textbf{HarMeme}} & \multicolumn{2}{c|}{\textbf{MAMI}} & \multicolumn{2}{c|}{\textbf{Harm-P}} & \multicolumn{2}{c|}{\textbf{MultiOFF}} & \multicolumn{2}{c}{\textbf{PrideMM}}  \\
 & Model  & \textbf{AUC} &  \textbf{Acc.} & \textbf{AUC} & \textbf{Acc.} & \textbf{AUC} &  \textbf{Acc.} & \textbf{Acc.} & \textbf{F1} & \textbf{Acc.} & \textbf{F1} & \textbf{Acc.} & \textbf{F1} \\ 
\midrule
\multicolumn{2}{c|}{\multirow{2}{*}{Best prior results}}& \multicolumn{2}{c|}{VPD-55B} & \multicolumn{2}{c|}{ISSUES} & \multicolumn{2}{c|}{Pro-Cap} & \multicolumn{2}{c|}{ExplainHM} & \multicolumn{2}{c|}{RGCL} & \multicolumn{2}{c}{MemeCLIP} \\
\multicolumn{2}{c|}{}& 89.20 & 80.80 & 92.83 & 81.64 & 83.77 & 73.63 & \underline{90.7} & \underline{90.7} & 67.13 & 58.11 & 76.06 & 75.09 \\
 \midrule
 \multicolumn{8}{l}{\textit{~~~~~Supervised fine-tuned CLIP-based Classifiers}} \\
%Underline the second best
 \midrule
1& CLIP  & 79.81 & 72.04 & 82.63 & 76.78 & 77.66 & 68.44 & 80.55 & 80.25 & 62.41 & 48.14 & 72.39 & 72.33   \\
%PromptHate & 81.5 & 73.0 & 90.9 & 84.5\\ 
%ISSUES &  85.51 & 77.70  &  92.83 & 81.64 \\
%DisMultiHate & 82.8 & 75.8 & 86.39 & 81.24 & & & & \\
%2& MOMENTA & 69.17 & 61.34 & 86.32 & 80.48 & 81.68 & 72.10 & 89.84 & 88.26 & \text{-} & \text{-} & 72.23 & 71.78  \\ 
2& HateCLIPper & 85.48 & 76.10 & 89.72 & 84.80 & 87.15 & 74.82 & 87.60 & 86.90 & 62.40 & 54.80 & 75.53 & 74.08  \\
3& RGCL & 87.04 & 78.82 & 91.81 & 87.03 & 89.39 & 78.35 & 89.92 & 89.51 & 67.13 & 58.11  & 76.34 & 76.50 \\ 
\midrule
\multicolumn{8}{l}{\textit{~~~~~Large Multimodal Models}} \\\midrule

4 & GPT-4o & \text{-} & 71.3 & \text{-} & 72.9 & \text{-} & 79.4 & 63.1 & 64.5 & 58.3 & 58.1 & 75.3 & 73.7  \\
\multicolumn{2}{c|}{LLaVA-1.5-7B} & \multicolumn{2}{c|}{} & \multicolumn{2}{c|}{} & \multicolumn{2}{c|}{}& \multicolumn{2}{c|}{}& \multicolumn{2}{c|}{} \\
5 & ~\textit{w/ zero-shot} & 63.69 & 57.60 & 71.38 & 48.59 & 67.60 & 58.30  &  61.57  & 46.39 & 59.55 & 51.68 & 63.43 & 65.64 \\
6 & ~\textit{w/ few-shot} & 63.38 & 57.20 & 73.42 & 59.60 & 68.09 & 62.70 & 53.52 & 52.17 & 38.93 & 56.04 & 62.13 & 64.04
  \\
7 & ~\textit{w/ SFT} & 85.19 & 78.71 & 91.38 & 79.10 & 85.97 & 73.92 & 82.76 & 82.76 & 67.79 & 57.81 & 73.18 & 75.97 \\
8 & ~\textit{w/ RA-HMD} & \underline{89.7} & \underline{80.9} & \textbf{93.5} & \textbf{88.2} & \textbf{91.2} & \underline{79.7} & 89.64 & 89.34 & \underline{70.9} & \underline{63.6} & \textbf{78.1} & \textbf{78.7} \\
%& ~\textit{p}-value & \multicolumn{1}{c}{9.8$e^{-3}$} &  \multicolumn{1}{c|}{3.5$e^{-3}$} & \multicolumn{1}{c}{1.2$e^{-2}$} & \multicolumn{1}{c|}{8.5$e^{-3}$} & \multicolumn{1}{c}{4.4$e^{-3}$} & \multicolumn{1}{c|}{6.2$e^{-3}$} & \multicolumn{1}{c}{2.5$e^{-3}$} & \multicolumn{1}{c|}{1.6$e^{-3}$} & \multicolumn{1}{c}{6.1$e^{-3}$} & \multicolumn{1}{c|}{4.6$e^{-3}$} & \multicolumn{1}{c}{5.6$e^{-3}$} & \multicolumn{1}{c}{8.9$e^{-3}$} \\ 
 \multicolumn{2}{c|}{Qwen2-VL-2B} & \multicolumn{2}{c|}{} & \multicolumn{2}{c|}{} & \multicolumn{2}{c|}{}& \multicolumn{2}{c|}{}& \multicolumn{2}{c|}{} \\
9 & ~\textit{w/ zero-shot} & 64.84 & 54.20 & 61.05 & 56.67 & 67.19 & 51.01 & 53.92 & 21.78 & 63.11 & 36.33 & 57.79 & 53.27 \\
10 & ~\textit{w/ few-shot} & 61.72 & 59.10 & 62.11 & 65.82 & 64.84 & 58.80 & 52.96 & 51.62 & 67.11 & 44.92 & 55.42 & 54.30  \\
11 & ~\textit{w/ SFT} & 83.98 & 76.20 & 90.23 & 82.49 & 77.73 & 68.64 & 80.25 & 79.69 & 66.44 & 54.53 & 73.74 & 74.17 \\
12 & ~\textit{w/ RA-HMD} & 88.41 & 79.13 & 92.93 & 87.66 & 89.26 & 79.40 & 88.91 & 88.70 & 68.46 & 61.79 & 76.03 & 76.72 \\
%& ~\textit{p}-value & \multicolumn{1}{c}{4.2$e^{-3}$} &  \multicolumn{1}{c|}{4.8$e^{-3}$} & \multicolumn{1}{c}{9.1$e^{-3}$} & \multicolumn{1}{c|}{7.6$e^{-3}$} & \multicolumn{1}{c}{6.5$e^{-4}$} & \multicolumn{1}{c|}{1.9$e^{-4}$} & \multicolumn{1}{c}{9.3$e^{-4}$} & \multicolumn{1}{c|}{1.1$e^{-3}$} & \multicolumn{1}{c}{2.0$e^{-2}$} & \multicolumn{1}{c|}{7.7$e^{-3}$} & \multicolumn{1}{c}{7.1$e^{-3}$} & \multicolumn{1}{c}{8.3$e^{-3}$} \\ 
 \multicolumn{2}{c|}{Qwen2-VL-7B} & \multicolumn{2}{c|}{} & \multicolumn{2}{c|}{} & \multicolumn{2}{c|}{}& \multicolumn{2}{c|}{}& \multicolumn{2}{c|}{} \\
13 & ~\textit{w/ zero-shot} & 71.88 & 63.20 & 64.84 & 64.12 & 76.17 & 58.50 & 55.49 & 22.85 & 63.44 & 35.94 & 65.29 & 62.89   \\
14 & ~\textit{w/ few-shot} & 71.48 & 63.80 & 71.48 & 67.23 & 73.44 & 66.10 & 55.55 & 65.23 & 64.43 & 54.69 & 69.14 & 56.61 \\
15 & ~\textit{w/ SFT} & 86.33 & 78.60 & 91.80 & 85.88 & 82.64 & 72.39 & 85.89 & 86.33 & 67.79 & 55.51 & 75.10 & 74.88    \\
16 & ~\textit{w/ RA-HMD} &
\textbf{91.1}
 & \textbf{82.1} & \underline{93.2} & \underline{88.1} & \underline{90.4} & \textbf{79.9} & \textbf{91.6} & \textbf{91.1} & \textbf{71.1} & \textbf{64.8} & \textbf{78.1} & \underline{78.4} \\ 

%& Average gain $\uparrow$& +4.56  & +2.87 & +2.07 & +5.49 & +8.17 & +8.04  &  +5.74 & +5.46 & +2.81 & +6.29 & +2.05 & +1.60\\  
%& ~\textit{p}-value & \multicolumn{1}{c}{8.7$e^{-4}$} & \multicolumn{1}{c|}{2.4$e^{-3}$} & \multicolumn{1}{c}{3.5$e^{-2}$} & \multicolumn{1}{c|}{1.6$e^{-2}$} & \multicolumn{1}{c}{2.5$e^{-3}$} & \multicolumn{1}{c|}{2.6$e^{-3}$} & \multicolumn{1}{c}{6.3$e^{-3}$} & \multicolumn{1}{c|}{8.6$e^{-3}$} & \multicolumn{1}{c}{1.3$e^{-2}$} & \multicolumn{1}{c|}{5.9$e^{-3}$} &\multicolumn{1}{c}{9.2$e^{-3}$} & \multicolumn{1}{c}{7.2$e^{-3}$}\\

\bottomrule
\end{tabular}
}}
\caption{Comparison with baseline systems under supervised settings. For large multimodal models, we report the pre-trained models zero-shot and few-shot performance (using 4-shot evaluation), along with a comparison between SFT and RA-HMD. 
%Additionally, for each LMM, we provide the ~\textit{p}-value from significance testing between SFT and RA-HMD. 
%All ~\textit{p}-values are below 0.05. %\howard{For pre-trained LMM, we use the LMH for inference. For RA-HMD, we use LRC.} 
Best performance is highlighted in \textbf{bold}; second-best is \underline{underlined}.}
\label{tab:results_supervised}
\end{table*}

\subsection{Comparing RA-HMD to Baseline Systems under Supervised Settings}
\label{sec:results_supervise}
Table~\ref{tab:results_supervised} presents the performance of baseline systems under supervised fine-tuning settings. We compare RA-HMD against a range of strong baselines: the best prior models for each dataset\footnote{From a recent paper \cite{Nguyen_2024_computationalMemeUnderstanding}; some datasets have been updated with the new best results.}; supervised fine-tuned CLIP-based classifiers; and Large Multimodal Models (LMMs). 
All models are fine-tuned and evaluated for each dataset separately. 

\textbf{CLIP-based Classifiers}
We compare the performance of fine-tuned CLIP \cite{clip2021} model with two other fine-tuning methods for CLIP-based systems: HateCLIPper \cite{KumarHateClip2022} and RGCL \cite{RGCL2024Mei}.

\textbf{Large Multimodal Models}
We experiment with three LMMs from two model families: LLaVA-1.5-7B \cite{Liu_2023_LLAVA1.5}, Qwen2-VL-2B and Qwen2-VL-7B \cite{Wang_2024_Qwen2VL}. We report the performance of these LMMs in the following settings: pre-trained models with zero-shot and few-shot prompts using the LMH; SFT LMMs using the LMH; and classification using LRC under the RA-HMD fine-tuning framework. We further include the results with GPT-4o \cite{OpenAI2024gpt4o} with optimized prompting for each dataset for reference. For GPT-4o, the token likelihood is not accessible to compute the AUC score.

\textbf{Best Prior Models}
 Visual Program Distillation (VPD) \cite{Hu_2024_VPD} and ExplainHM \cite{Lin_2024_ExplainHM} are LLM agent-based systems. 
 %VPD fine-tunes PaLI-X 55B \cite{Chen2023PALIX} to utilize tools and execute programs, while ExplainHM fine-tunes three LLMs as two debaters and one judge to explain and classify hateful memes. 
 The remaining state-of-the-art models, including ISSUES \cite{Burbi_2023_Issues}, Pro-Cap \cite{Cao_2023_ProCap}, RGCL \cite{RGCL2024Mei} and MemeCLIP \cite{Shah2024memeclip_pridemm}, are based on fine-tuning CLIP-based vision and language models. Detailed descriptions of these methods are provided in Appendix~\ref{appendix:baseline_models}.

\begin{table*}[h]
\small
\sisetup{table-format=2.1, table-auto-round=true, table-number-alignment=center, detect-weight=true, detect-family=true,mode=text}
\centering
\setlength{\tabcolsep}{5pt}{
\resizebox{1.0\linewidth}{!}{%
\begin{tabular}{ll|SS| SS| SS| SS| SS| SS}
\toprule 
&\multicolumn{1}{l|}{\textbf{Evaluated on}}&
 \multicolumn{2}{c|}{\textbf{HatefulMemes}} & \multicolumn{2}{c|}{\textbf{HarMeme}} & \multicolumn{2}{c|}{\textbf{MAMI}} & \multicolumn{2}{c|}{\textbf{Harm-P}} & \multicolumn{2}{c|}{\textbf{MultiOFF}} & \multicolumn{2}{c}{\textbf{PrideMM}}  \\
 &Model  & \textbf{AUC} &  \textbf{Acc.} & \textbf{AUC} & \textbf{Acc.} & \textbf{AUC} &  \textbf{Acc.} & \textbf{Acc.} & \textbf{F1} & \textbf{Acc.} & \textbf{F1} & \textbf{Acc.} & \textbf{F1} \\ 
\midrule
\multicolumn{8}{l}{\textit{~~~~~Low resourced systems}}\\\midrule
1 & GPT-4o & \text{-} & 66.40 & \text{-}  & 68.4 & \text{-} & 72.90 & 55.39 & 55.11 & 61.07 & 51.07 & 63.79 & 62.28   \\ % add to both tables 
2 & Mod-Hate & 64.5 & 58.0 & 73.4 & 69.5&  67.4 & 61.0 & \text{-} &  \text{-} & \text{-} & \text{-} & \text{-} & \text{-}\\ 
3 & LOREHM &  \text{-} & 65.6 & \text{-}  & 73.73 & \text{-} & \underline{75.4} & \text{-} &  \text{-} & \text{-} & \text{-} & \text{-} & \text{-} \\

\midrule 
%&        63.69 & 57.60 & 71.38 &   48.59 & 46.39 & 61.57 & 51.68 & 59.55 & 63.43 &  65.64 & 58.30 & 67.60 \\
\multicolumn{8}{l}{\textit{~~~~~Systems fine-tuned under cross-dataset settings}}\\\midrule
& \multicolumn{1}{l|}{Fine-tuning set }&
 \multicolumn{2}{c|}{ HarMeme} & \multicolumn{10}{c}{HatefulMemes}  \\
 \midrule
4 & RGCL & 69.9 &\underline{66.9} & 64.32 & 61.11 & 67.8 & 62.4& 56.42 & 57.10 & 53.69 & 45.1 & 59.76 & 61.48   \\ 
 & LLaVA-1.5-7B & \multicolumn{2}{c|}{} & \multicolumn{2}{c|}{} & \multicolumn{2}{c|}{}& \multicolumn{2}{c|}{}& \multicolumn{2}{c|}{} \\
5 & ~~~\textit{SFT + zero-shot}
& 63.79 & 59.40 & 61.92 & 48.47 & 69.14 & 61.10 & 55.21 & 28.70 & 62.76 & 32.50 & 58.11 & 53.25 \\ 
6 & ~~~\textit{SFT + few-shot} & 63.07 & 56.40 & 69.88 & 52.82 & 65.54 & 50.10 & 55.64 & 49.58 &  56.04 & 38.93 & 48.52 & 55.34 \\ 
7 & ~~~\textit{RA-HMD + RKC} & \underline{74.2} & 65.17 & \textbf{89.5} & \textbf{81.9} & \underline{80.0} & 74.45 & \textbf{ 67.3} & \textbf{67.8} & 62.42 & 51.73 & 68.84 & 67.72 \\ 
 & Qwen2-VL-2B & \multicolumn{2}{c|}{} & \multicolumn{2}{c|}{} & \multicolumn{2}{c|}{}& \multicolumn{2}{c|}{}& \multicolumn{2}{c|}{} \\
8 & ~~~\textit{SFT + zero-shot}
& 64.06 & 59.70 & 61.30 & 52.16 & 66.41 & 57.30 & 53.52 & 20.31 & 62.33 & 29.30 & 56.41 & 58.98 \\ 
9 & ~~~\textit{SFT + few-shot} & 61.33 & 53.80 & 57.42 & 64.97 & 73.83 & 66.00 & 56.89 & 55.49 & 53.69 & 42.42 & 55.42 & 60.94 \\ 
10 & ~~~\textit{RA-HMD + RKC} & 70.86 & 62.77 & 85.98 & 78.37 & 74.79 & 72.27 & 63.38 & 65.97 & \underline{63.4} & 53.37 & \underline{69.0} & \underline{69.1} \\ 
 & Qwen2-VL-7B & \multicolumn{2}{c|}{} & \multicolumn{2}{c|}{} & \multicolumn{2}{c|}{}& \multicolumn{2}{c|}{}& \multicolumn{2}{c|}{}\\
11 & ~~~\textit{SFT + zero-shot} & 71.09 & 64.10 & 62.99 & 55.16 & 71.09 & 61.90 & 54.65 & 21.48 & 63.13 & 29.69 & 64.47 & 63.58 \\ 
12 & ~~~\textit{SFT + few-shot} & 72.27 & 60.60 & 67.19 & 62.43 & 73.44 & 66.00 & 56.36 & 64.90 & 62.02 & \underline{53.7} & 55.42 & 60.94 \\ 
13 & ~~~\textit{RA-HMD + RKC} & \textbf{77.1} & \textbf{69.3} & \underline{88.8} & \underline{81.7} & \textbf{81.4} & \textbf{75.6} & \underline{64.5} & \underline{66.4} & \textbf{63.8} & \textbf{55.6} & \textbf{69.3} & \textbf{69.3} \\ 
\bottomrule
\end{tabular}
}}
\caption{Comparing out-of-domain meme classification performance under low-resource settings. For systems fine-tuned under cross-dataset settings, models are fine-tuned on HarMeme and evaluated on the HatefulMemes dataset. For the remaining evaluation datasets, models are fine-tuned on the HatefulMemes dataset. For LMMs, we compare the SFT models using zero-shot and in-context learning with the RA-HMD fine-tuned models using RKC. Few-shot examples (4-shot) and RKC examples are drawn from the training split of each evaluation dataset. \#2 and \#3 are taken from the original paper. 
Best performance is highlighted in \textbf{bold}; second-best is \underline{underlined}. }
\label{tab:results_low_resource}
\end{table*}
\textbf{Observation 1: Fine-tuned CLIP-based classifiers outperform baseline LMMs.}\\
As shown in Table~\ref{tab:results_supervised}, RGCL (\#3) achieves the highest performance among CLIP-based classifiers, surpassing standard fine-tuned CLIP (\#1) by approximately 10\% across multiple datasets. On 5 out of 6 datasets, RGCL performs better than, or on par with, all three SFT LMMs (\#7, \#11, \#15).
%These results align with \citealp{RGCL2024Mei}'s finding and underpin that we need better fine-tuning approaches for LMMs. The two exceptions are MultiOFF and PrideMM. On MultiOFF, Qwen2-VL-7B SFT performs slightly better on accuracy with 0.7\% gains over RGCL, however, its F1 score is almost 3\% lower than RGCL. On PrideMM, RGCL performs slightly worse than Qwen2-VL-7B SFT.

\textbf{Observation 2: In-context learning exhibits limited efficacy for meme classification.}\\
We compare the zero-shot (\#5, \#9, \#13) and few-shot (\#6, \#10, \#14) performance of the pre-trained LMMs. Our findings indicate that, in-context learning does not benefit meme classification, which is consistent with previous results \cite{Hee2024BridgeModality, Huang_LowResourceLMMAgentHatefulMeme_2024}. HarMeme is the only dataset where few-shot systems consistently outperform zero-shot systems. On Harm-P and MultiOFF, although the accuracies of zero-shot and few-shot remain comparable, the few-shot experiments yield a significant gain in F1 score. This improvement is due to a more balanced precision and recall after providing demonstration examples to the system.
% Precision and Recall balance explanation in appendix
% Try find explanation for harmeme's gain 

\textbf{Observation 3: RA-HMD outperforms all strong baseline systems across six datasets}\\
Across six datasets and three LMMs, fine-tuning with RA-HMD significantly improves performance over SFT (Table~\ref{tab:results_supervised}: \#7, \#8; \#11, \#12; \#15, \#16). Statistical significance tests comparing RA-HMD and SFT further validate these results, with all p-values below 0.05 (see Appendix~\ref{appendix:statistical_sig}). Notably, as indicated in \#16, Qwen2-VL-7B fine-tuned with RA-HMD outperforms VPD-PaLI-X-55B on HatefulMemes. Moreover, RA-HMD improves upon RGCL with gains of over 4\% in AUC and 3\% in accuracy on HatefulMemes. These gains show RA-HMD’s effectiveness in improving LMMs for meme classification over SFT.

\subsection{Comparing RA-HMD with Baseline Systems under Low-Resource Settings}
\label{sec:results_lowres}
Online hate speech is constantly evolving, posing a challenge to systems as the distribution of memes encountered in the wild departs from that of the training data. To simulate real-world deployment constraints, we evaluate systems on out-of-domain examples under low-resource settings where gradient updates are prohibited and only demonstration examples are available \cite{Huang_LowResourceLMMAgentHatefulMeme_2024, Hee2024BridgeModality, Cao_2024_ModHate}.

We adopt a cross-dataset evaluation protocol similar to \citet{RGCL2024Mei}: models fine-tuned on \textbf{HarMeme} are evaluated on \textbf{HatefulMemes}, while models trained on \textbf{HatefulMemes} are evaluated on all other datasets. This protocol simulates a scenario in which a trained meme classification system is deployed to evaluate trending memes. Few-shot and RKC examples are drawn from the training split of each of the target evaluation datasets to avoid test set contamination.

%We compare our Retrieval-augmented KNN Classifier (RKC) fine-tuned with RA-HMD against the following systems: 
We compare RA-HMD fine-tuned LMM with the RKC against the following systems: SFT LMMs with zero-shot and few-shot prompting using LMH; GPT-4o \cite{OpenAI2024gpt4o}; specialized low-resource systems (LOREHM \cite{Huang_LowResourceLMMAgentHatefulMeme_2024}, Mod-hate \cite{Cao_2024_ModHate}). For GPT-4o, we report results without prompt optimization for each dataset, as this setting assumes the hate type is not known in advance. Further discussion and comparison of GPT-4o results can be found in Appendix~\ref{appendix:gpt4o}.

\textbf{Observation 1: Fine-tuning on one memes classification dataset does not help LMMs to improve generalization on other meme classification datasets}\\
Cross-domain fine-tuned LMMs show no consistent improvements over pre-trained LMMs for either zero-shot or few-shot prompting.  Qwen2-VL-7B zero-shot (\#11 in Table~\ref{tab:results_low_resource}) matches its SFT model performance (\#13 in Table~\ref{tab:results_supervised}) on HatefulMemes and PrideMM but has performance degradation on the remaining four datasets.

\textbf{Observation 2:  SFT LMMs with in-context learning is ineffective}\\
%Consistent with Observation 2 of Section~\ref{sec:results_supervise}, after LMMs are fine-tuned on different domains of hateful meme datasets, the few-shot approach remains similarly ineffective as shown in Table~\ref{tab:results_generalize} \#6, \#9 \#12.
As shown in Table~\ref{tab:results_low_resource} \#6, \#9 and \#12, the few-shot approach remains similarly ineffective after LMMs are fine-tuned on different domains of hateful meme datasets, offering no significant gains over the SFT zero-shot models (Table~\ref{tab:results_low_resource} \#5, \#8,  \#11).
%The same conclusions apply, highlighting the limited benefit of in-context learning in these scenarios. 
In Section~\ref{sec:ablate_icl_rkc}, we further analyze the effect of the number of shots in in-context learning and find that increasing the number of shots does not improve performance. Moreover, in Section~\ref{sec:abl_RKC_ICL_Models}, we show that in-context learning with RA-HMD consistently outperforms SFT models, demonstrating RA-HMD’s effectiveness in enhancing out-of-domain generalization.

\textbf{Observation 3: RA-HMD fine-tuned LMMs with RKC inference mode outperforms baseline methods}\\
RA-HMD fine-tuned LMMs using RKC outperform the baseline SFT LMMs in both zero-shot and few-shot settings under the same cross-dataset fine-tuning settings. Notably, RA-HMD trained Qwen2-VL-7B with RKC improves over the baseline SFT few-shot model by 21.6\% in AUC and 19.3\% in accuracy on HarMeme (Table~\ref{tab:results_low_resource} \#11-13). The ablation study in Section~\ref{sec:ablate_icl_rkc}, which varies the number of top k for RKC, further demonstrates that RKC uses demonstration examples more effectively than the few-shot in-context learning framework. Moreover, as shown in Appendix~\ref{appendix:rkc_icl_ft}, applying RKC to both pretrained and SFT LMMs results in worse performance compared to RA-HMD, underscoring the effectiveness of our fine-tuning strategy.

\textbf{Observation 4: RA-HMD trained LMMs with RKC inference outperform other low resource methods}\\
%Compared to GPT-4o, our RA-HMD fine-tuned Qwen2-VL-7B with RKC achieves 14.5\% higher accuracy on HarMeme and 11.3\% higher accuracy on PrideMM. 
When compared to other low-resource methods, our RA-HMD fine-tuned LLaVA-1.5-7B with RKC matches the performance of LOREHM on the HatefulMemes dataset. Notably, LOREHM uses a newer and larger LLaVA-1.6-34B within an agent-based framework. Furthermore, our method outperforms LOREHM by 8.2\% in accuracy on HarMeme, highlighting our methods' effectiveness under low-resource settings.

% two observations baseline/gpt4o
%\subsection{Ablation Study}
\subsection{Effects of Two-Stage Fine-tuning}
We assess the contribution of each stage within our two-stage RA-HMD fine-tuning process. As shown in Tables~\ref{tab:ablation_stages}, omitting either stage leads to performance degradation in both supervised and cross-dataset settings, with Stage 1 contributing more substantial gains.

When only Stage 1 is applied and Stage 2 is omitted, the performance loss is less severe under supervised settings than in cross-dataset evaluations. 
We attribute this to the contrastive loss in Stage 2, which explicitly optimizes retrieval by aligning representations of semantically similar meme pairs, thereby enhancing robustness to distribution shifts in unseen datasets.

We also compared a variant where we jointly optimize the losses from both stages in a single training phase.
\begin{equation}
\mathcal{L}_i^{\textrm{Combined}} = \mathcal{L}_i^{CL} + \mathcal{L}_i^{LR} + \mathcal{L}_i^{LM}.
\label{eq:combined}
\end{equation}
This combined training yields suboptimal results, demonstrating that the two-stage fine-tuning effectively resolves the optimization conflict between task adaptation (Stage 1) and representation alignment (Stage 2). Furthermore, since the LMM remains trainable throughout combined training, updating the encoded meme database incurs significantly higher computational costs compared to the two-stage fine-tuning approach, where the LMM is frozen in stage 2. This staged separation thus enables more efficient training while obtaining stronger performance. %Additional details on training time are provided in Appendix~\ref{appendix:run_time}.

%We also conduct ablation studies by removing individual loss terms within each stage. Results show that every loss component is essential for optimal performance. Full details are shown in Appendix~\ref{appendix:abl_loss}.
In Appendix~\ref{appendix:abl_loss}, we also conduct ablation studies by removing individual loss terms within each stage and found that every loss component is essential.
\begin{table}[htb]
\centering

\begin{subtable}[t]{0.47\textwidth}
\small
\centering
\begin{tabular}{l|ll|ll}
\toprule
 &\multicolumn{2}{c|}{\textbf{HatefulMemes}} & \multicolumn{2}{c}{\textbf{HarMeme}}\\
 Mode & \textbf{AUC} & \textbf{Acc.} & \textbf{AUC} & \textbf{Acc.} \\ \midrule

%SFT &  86.3 & 78.6 & 91.8 & 85.9 \\
RA-HMD & \textbf{91.1} & \textbf{82.1} & \textbf{93.2} & \textbf{88.1} \\
~\textit{w/ Stage 1 only} & 90.2 & 81.4 & 92.0 & 86.2 \\
~\textit{w/ Stage 2 only} & 84.4 & 74.2 & 90.1 & 85.6 \\
~\textit{w/ $\mathcal{L}_i^{\textrm{Combined}}$} & 88.9 & 77.8 & 90.2 & 83.4 \\
\bottomrule
\end{tabular}
\caption{Supervised settings, see Table~\ref{tab:results_supervised} for detailed settings}
\label{tab:ablation_stages_supervised}
\vspace{3pt}
\end{subtable}
\hfill
\begin{subtable}[t]{0.47\textwidth}
\small
\centering
\begin{tabular}{l|ll|ll}
\toprule
 &\multicolumn{2}{c|}{\textbf{HatefulMemes}} & \multicolumn{2}{c}{\textbf{HarMeme}}\\
 Mode & \textbf{AUC} & \textbf{Acc.} & \textbf{AUC} & \textbf{Acc.} \\ \midrule
%SFT & 71.1 & 64.1 & 63.0 & 55.2 \\
RA-HMD & \textbf{77.1} & \textbf{69.3} & \textbf{88.8} & \textbf{81.7} \\ 
~\textit{w/ Stage 1 only} & 74.4 & 66.7 & 86.3 & 78.7 \\
~\textit{w/ Stage 2 only} & 72.0 & 62.1 & 84.9 & 78.1 \\
~\textit{w/ $\mathcal{L}_i^{\textrm{Combined}}$} & 72.2 & 65.3 & 87.5 & 80.2 \\
\bottomrule
\end{tabular}
\caption{Cross-dataset settings, see Table~\ref{tab:results_low_resource} for detailed settings}
\label{tab:ablation_stages_cross}
\end{subtable}
\caption{Ablation study of RA-HMD two-stage fine-tuning framework on Qwen2-VL-7B, evaluating the impact of Stage 1 and Stage 2 Fine-tuning. For $\mathcal{L}_i^{\textrm{Combined}}$, we jointly optimize the three loss objectives from both stages in a single training process as shown in Eq.~\ref{eq:combined}.}
\label{tab:ablation_stages}
\end{table}

\subsection{Performance on General Vision-Language Benchmarks}
\label{sec:general}
Table~\ref{tab:generalVisual} compares the pretrained Qwen2-VL-2B with its SFT and RA-HMD variants, both fine-tuned on the HatefulMemes dataset, across three general vision-language benchmarks: MMMU \cite{Yue2023MMMU}, SEED-Bench \cite{Li2023SeedBench}, and GQA \cite{GQA2019}. Evaluation settings and additional results are provided in Appendix~\ref{appendix:general_VL}. The SFT model shows performance degradation across all three benchmarks, while RA-HMD maintains performance comparable to the pretrained model. These results indicate that RA-HMD robustly preserves the general vision-language capabilities of LMMs.
\begin{table}[h]
\small
\centering
\begin{tabular}{llll}
\toprule
Model & MMMU & SEEDBench & GQA \\
\midrule
Qwen2-VL-2B & 40.2 & 72.7 & 60.4 \\
~\textit{+SFT} & 39.1 & 72.1 & 57.0 \\
~\textit{+RA-HMD} & 40.4 & 72.7 & 60.1\\
\bottomrule
\end{tabular}
\caption{Comparison of the pretrained, SFT, and RA-HMD Qwen2-VL-2B models on three general vision-language benchmarks. The SFT and RA-HMD models are fine-tuned on the HatefulMemes dataset.}
\label{tab:generalVisual}
\end{table}

\subsection{Robustness Under Adversarial Attack}
We follow \citealt{Hateproof2023} to assess the system's robustness under adversarial attack. Specifically, we adopt the SaltPepper-I-High attack described in their work, which injects white and black pixels across the image in a manner that does not compromise the overall perception of semantic content.

We compare the SFT and the RA-HMD tuned Qwen2-VL-7B systems (corresponding to \#15 and \#16 in Table~\ref{tab:results_supervised}). As shown in Table~\ref{tab:adversarial_robustness}, RA-HMD consistently outperforms SFT on SaltPepper-I-High–perturbed data, exhibiting less severe performance degradation. Moreover, when the perturbed examples are incorporated into the retrieval database alongside the original ones, robustness improves further. This shows that our RA-HMD retrieval-guided approach offers a simple yet effective way to enhance system robustness against adversarial attacks as soon as these attacks are detected.
\begin{table}[t]
\small
\centering
\begin{tabular}{l|ll}
\toprule
 & \multicolumn{2}{c}{\textbf{HatefulMemes}} \\
 Model & \textbf{AUC} & \textbf{Acc.} \\ \midrule
\multicolumn{3}{l}{~Baseline (From Table~\ref{tab:results_supervised})} \\
\midrule
SFT & 86.3 & 78.6 \\
RA-HMD + \textit{LRC} & 91.1 & 82.1 \\
RA-HMD + \textit{RKC} & 90.8 & 81.8 \\
\midrule
\multicolumn{3}{l}{~Under Adversarial Attack} \\
\midrule
SFT & 80.5 (-5.8) & 72.3 (-6.3) \\
RA-HMD + \textit{LRC} & 84.4 (-6.7) & 75.5 (-6.6) \\
RA-HMD + \textit{RKC} & 86.8 (-4.0) & 76.6 (-5.2) \\
~\textit{w/ Augmented DB} & \textbf{88.4} (-2.4) & \textbf{78.4} (-3.4) \\
\bottomrule
\end{tabular}
\caption{Comparison of the SFT and RA-HMD Qwen2-VL-7B models on HatefulMemes under adversarial attack. Values in parentheses denote performance drop compared to non-attack.}
\label{tab:adversarial_robustness}
\end{table}

\subsection{Numbers of Shots and Neighbors}
\label{sec:ablate_icl_rkc}
In Appendix~\ref{appendix:ablate_icl_rkc}, we ablate the effects of varying the number of shots for few-shot in-context learning and varying the number of top K nearest neighbors for RKC. %Due to space constraints, the full results are provided in Appendix~\ref{appendix:ablate_icl_rkc}. 
We find that merely adding more shots does not necessarily improve performance for in-context learning, aligning with the findings of \citet{Huang_LowResourceLMMAgentHatefulMeme_2024}.  In contrast, increasing $K$ in RKC leads to steady performance gains, with improvements plateauing around $K=20$. These results suggest that the RKC inference mode of RA-HMD makes more effective use of demonstration examples compared to in-context learning.

\subsection{Comparing Out-of-Domain Generalization Across Model Variants}
\label{sec:abl_RKC_ICL_Models}
To further evaluate out-of-domain generalization across different model variants, we compare RA-HMD against both pretrained and SFT models in Appendix~\ref{appendix:rkc_icl_ft}, following the same protocol as in Table~\ref{tab:results_low_resource}.
Our results show that RA-HMD consistently outperforms both baselines under both the in-context learning framework and the retrieval-augmented KNN classifier (RKC), highlighting its robustness and effectiveness for generalizing to unseen meme distributions.

\subsection{Comparing Different Inference Modes}
\label{sec:abl_infer_mode}
%In our experiments the default inference mode varies by setting: for the pre-trained and SFT models, we use the \howard{LM Head} (LMH), whereas for RA-HMD under supervised settings, we use logistic regression (LR). The Retrieval-augmented KNN Classifier (RKC) is employed as the default for RA-HMD in cross-dataset settings. 
We compare RA-HMD using three classifiers: LMH, LRC and RKC in Appendix~\ref{appendix:abl_infer_mode}. Under supervised settings, the performance is similar. However, in cross-dataset scenarios, RKC outperforms both LMH and LRC, underscoring its superior effectiveness in handling out-of-domain examples.
\subsection{Comparing Rationales Generated by Models}
We compare meme explanations generated by the SFT and RA-HMD fine-tuned Qwen2-VL-7B models (Table~\ref{tab:results_supervised} rows 15 and 16) on the validation set of the HatefulMemes dataset, where human-annotated rationales are available as ground-truth references \cite{Hee_DecodeMeaningFBHM_Hatred_2023}. RA-HMD is evaluated after Stage-1 fine-tuning, since Stage-2 does not fine-tune the Language Model Head for language generation.

Following prior work \cite{Yang2023Hare}, we evaluate explanation quality using two approaches based on LLM-as-judge:
\begin{itemize}
    \item pair-wise comparison,
    \item rubric-based evaluation.
\end{itemize}
The LLM judge measures how closely model-generated explanations align with human rationales. Full details of the evaluation protocol are provided in Appendix~\ref{appendix:rationale}.

The pair-wise comparison results are:
\begin{itemize}
    \item RA-HMD beats SFT: 61.5% 
    \item RA-HMD ties SFT: 13.8%
    \item SFT beats RA-HMD: 24.7%
\end{itemize}
and these results are visualized in Figure~\ref{fig:pairwise_compare}.

For the rubric-based evaluation (scored on a scale of 0–10), the SFT baseline achieved an average score of 4.9, while RA-HMD obtained a higher score of 5.6, further indicating stronger alignment with human rationales.

Based on the analysis of the generated explanations, we find that improvements in classification accuracy are supported by a deeper semantic understanding of memes. For challenging examples, where comprehension of background events or fine-grained details from the image is required, the RA–HMD fine-tuned system generates explanations that are both more accurate and more informative. Nevertheless, with an average score of 5.6, the performance remains far from perfect.  Further discussion of this aspect is provided in Limitations.

\subsection{Demonstration examples}
In Appendix~\ref{appendix:case_analysis}, we present a case analysis comparing the classification results of RA-HMD and SFT. Appendix~\ref{appendix:rationale} provides example rationales generated by each model.
\section{Conclusion}
We propose RA-HMD, a robust adaptation framework for LMMs tailored for hateful meme classification. Our approach effectively improves both in-domain accuracy and out-of-domain generalization, achieving state-of-the-art results across six meme classification datasets while preserving the general vision-language capabilities of the underlying models. 

\section*{Limitations}

%Hate speech can be defined by different terminologies, such as online harassment, online aggression, cyberbullying, or harmful speech. The United Nations Strategy and Plan of Action on Hate Speech notes that definitions of hateful speech could be controversial and disputed \cite{united_nations_2020}. 
%According to the UK Online Harms White Paper, harms may also be insufficiently defined \cite{UKparliament2022}. 

\textbf{Insufficient Definition of Hate Speech:} Hate speech is described using various terminologies, including online harassment, online aggression, cyberbullying, and harmful speech. The United Nations Strategy and Plan of Action on Hate Speech acknowledges that definitions of hate speech can be controversial and subject to debate \cite{united_nations_2020}. Similarly, the UK Online Harms White Paper highlights that certain harms may be insufficiently defined \cite{UKparliament2022}.
\\
\textbf{Variation in the Definition of Hate Speech:} We acknowledge that the definition of hate speech can be subjective and varies across different cultural and legal contexts. 
To this end, we evaluate our methods on six widely used meme classification datasets, allowing for generalization across different definitions of hate speech. As the discourse on defining hate speech evolves, we align our research with this ongoing process and plan to incorporate new datasets as they become available.
\\
\textbf{Fine-grained Vision Understanding:} In our error analysis, we find that the system is unable to recognize subtle visual details in memes. Enhancing image understanding through a more powerful vision encoder could further improve performance, which we leave for future work.
%This limitation might arise from the models' inability to comprehend facial expressions, which remains a constraint of our approach. Such challenges could potentially be addressed with a more advanced vision encoder.
\\
\textbf{Preference and RL-based Tuning Methods for Reasoning:}
\label{sec:limitation_reasoning}
In this work, our analysis and methodology primarily build on SFT, which may partly explain the limitations in reasoning quality. Future research could benefit from integrating reasoning capabilities from emerging models such as OpenAI-o1 \cite{OpenAI2024o1} and DeepSeek-R1 \cite{DeepSeekAI2025_r1}. A deeper investigation into preference-based and RL-based tuning methods, including DPO and GRPO, may prove valuable for incentivizing stronger reasoning abilities in LMMs.

\section*{Ethical Statement}
\paragraph{Reproducibility.} Detailed experimental setups, implementation specifics, and hyperparameter settings are provided in Appendix~\ref{appendix:exp_setup} to ensure reproducibility. The source code can be accessed from GitHub.

\paragraph{Usage of Datasets.}
The datasets used in this study—HatefulMemes, HarMeme, MAMI, Harm-P, MultiOFF, and PrideMM—were curated for research purposes to combat online hate speech. We strictly adhere to the terms of use established by the dataset authors.

\paragraph{Societal benefits.}
Hateful meme detection systems, like RA-HMD, can be used to automatically detect hateful content online, contributing significantly to reducing online hate speech. By reducing hate speech, fostering safer digital environments, and supporting human content moderators, these systems can make a significant impact on online communication and safety. We believe these benefits are both substantial and essential in the broader effort to create a more secure and respectful digital space.

\paragraph{Intended use.}
We intend to enforce strict access controls for model release. The model will be available only to researchers who agree to our terms of use, which explicitly state that the system is designed solely for the detection and prevention of hateful speech. Its use for any purposes that promote, condone, or encourage hate speech or harmful content is strictly prohibited.

\paragraph{Misuse Potential.} 
Although our system is not inherently designed to induce bias, training on existing datasets such as HatefulMemes may inadvertently propagate existing biases towards certain individuals, groups, or entities~\cite{PramanickMomenta2021}. To mitigate the risk of unfair moderation resulting from these dataset-induced biases, it is essential to incorporate human oversight into the moderation process if deployed.

\paragraph{Deployment consideration.}
Cultural differences and subjective topics introduce biases in moderating online hate speech. Expressions that may seem benign to some can be deeply offensive to others.
Our RKC inference mode relies on retrieving examples that generalize well across various domains, allowing the creation of multiple retrieval sets tailored to diverse cultural sensitivities without requiring retraining. However, before deploying such systems, it is crucial to carefully evaluate dataset annotations, particularly when addressing cultural differences and subjective interpretations. Key factors include data curation guidelines, potential annotator biases, and the inherently context-dependent definitions of hate speech. These considerations are essential to ensuring the system is deployed responsibly and effectively across varied cultural contexts.

\paragraph{Environmental Impact}
Training large-scale models is computationally intensive and contributes to global warming due to heavy GPU usage. However, our approach mitigates this issue by fine-tuning LMMs using quantized LoRA, a parameter-efficient method. As a result, our system can be trained in under four hours on a single consumer-grade GPU RTX 3090, costing less than 1 USD, significantly reducing both training time and computational cost compared to full-scale LMM fine-tuning. Furthermore, since our method generalizes across different domains without requiring retraining, it further minimizes computational overhead.

\begin{comment}

\section*{Acknowledgments}
Jingbiao Mei is supported by Cambridge Commonwealth, European and International Trust for the undertaking of the PhD in Engineering at the University of Cambridge.

Jinghong Chen is supported by the Warwick Postgraduate Studentship from Christ’s College and the Huawei Hisilicon Studentship for the undertaking of the PhD in Engineering at the University of Cambridge.

Weizhe Lin is supported by a Research Studentship funded by Toyota Motor Europe (RG92562(24020)) for the undertaking of the PhD in Engineering at the University of Cambridge.

Prof. Bill Byrne holds concurrent appointments as a Professor of Information Engineering at Cambridge University and as an Amazon Scholar.  This publication describes work performed at Cambridge University and is not associated with Amazon.

We would also like to thank all the reviewers for their knowledgeable reviews.

\end{comment}
\section*{Acknowledgments}
Jingbiao Mei is supported by Cambridge Commonwealth, European and International Trust for the undertaking of the PhD in Engineering at the University of Cambridge.

Jinghong Chen is supported by the Warwick Postgraduate Studentship from Christ’s College and the Huawei Hisilicon Studentship for the undertaking of the PhD in Engineering at the University of Cambridge.

Guangyu Yang is supported by Cambridge Commonwealth, European and International Trust for the undertaking of the PhD in Engineering at the University of Cambridge.

Weizhe Lin is supported by a Research Studentship funded by Toyota Motor Europe (RG92562(24020)) for the undertaking of the PhD in Engineering at the University of Cambridge.

Prof. Bill Byrne holds concurrent appointments as a Professor of Information Engineering at Cambridge University and as an Amazon Scholar.  This publication describes work performed at Cambridge University and is not associated with Amazon.

We would also like to thank all the reviewers for their knowledgeable reviews that helped us strengthen the contribution. 

% Entries for the entire Anthology, followed by custom entries
%\bibliography{anthology,custom}
%\bibliographystyle{acl_natbib}
\bibliography{custom}

\appendix
\label{sec:appendix}
\section{Dataset details and statistics}
\label{appendix:data_stats}

Table~\ref{tab:dataset_stats} shows the data split for our evaluation datasets. 

\paragraph{HatefulMemes} \cite{KielaFBHMC2020}
Released by Meta in 2020, HatefulMemes contains 12,000 memes annotated as hateful or benign by trained experts. This benchmark dataset synthesizes memes targeting religion, race, disability, and gender. It includes confounder examples where the benign memes are generated by altering either the image or text to challenge models’ ability in multimodal reasoning. 

\paragraph{HarMeme and Harm-P} 

HarMeme is a dataset containing approximately 3,000 memes centered on COVID-19 related political memes. A companion dataset, Harm-P \cite{PramanickMomenta2021}, contains around 3,000 memes related to US politics. Although the original HarMeme was later renamed Harm-C in subsequent work, we adhere to its original name following previous studies \cite{caoPromptHate2022}. In HarMeme, memes are annotated into three classes: very harmful, partially harmful, and harmless. Consistent with prior work \cite{caoPromptHate2022, PramanickMomenta2021}, we merge the very harmful and partially harmful categories into a single hateful class, while treating harmless memes as benign.

\paragraph{MAMI} \cite{Fersini_MAMI_2022}
The MAMI dataset focuses on detecting misogynistic memes sourced from various social media platforms, including Twitter and Reddit, as well as meme creation and sharing websites, and even anti-women websites and forums. It contains annotation for two tasks: (1) binary classification of misogyny and (2) categorization of misogyny types. In this work, we address the binary task of identifying whether a meme is misogynistic.

\paragraph{MultiOFF} \cite{suryawanshi-etal-2020-MultiOFF}
MultiOFF consists of memes gathered from Reddit, Facebook, Twitter, and Instagram, curated specifically for the detection of offensive content. Notably, the training set is extremely small, containing fewer than 500 meme examples. We use this dataset to evaluate the applicability of our methods under ultra low-resource conditions.

\paragraph{PrideMM} \cite{Shah2024memeclip_pridemm}
PrideMM contains LGBTQ+-themed memes annotated for four tasks: hate speech detection, hate target identification, topical stance classification, and humor detection. In this work, we use the hate speech classification annotations for hateful meme detection.

\begin{table}[h]
\small
\centering

\begin{tabular}{l|cc|cc}
\toprule
Datasets  & \multicolumn{2}{c}{Train}  & \multicolumn{2}{c}{Test}  \\
& \#Benign & \#Hate & \#Benign & \#Hate  \\
\midrule
HatefulMemes &   5450 & 3050 & 500 & 500      \\
HarMeme  & 1949 & 1064 & 230 & 124          \\ 
MAMI &  4500 & 4500 & 500 & 500 \\ 
Harm-P & 1534 & 1486 & 173 & 182 \\ 
MultiOFF & 258 & 187 & 58 & 91 \\ 
PrideMM & 2581 & 2482 & 260 & 247  \\ 

\bottomrule
\end{tabular}%
\caption{Statistical summary of HatefulMemes and HarMeme datasets}
\label{tab:dataset_stats}
\end{table}

For HatefulMemes, HarMeme, and MAMI, we report the Area Under the Receiver Operating Characteristic Curve (AUC) and Accuracy (Acc) in line with previous studies \cite{KumarHateClip2022, Cao_2023_ProCap, RGCL2024Mei, Cao_2024_ModHate}. For Harm-P, MultiOFF, and PrideMM, we report Accuracy and F1 score consistent with the literature \cite{PramanickMomenta2021, RGCL2024Mei, Shah2024memeclip_pridemm, Lin_2024_ExplainHM}.

To access the Facebook HatefulMemes dataset, one must follow the license from Facebook\footnote{\href{https://hatefulmemeschallenge.com/\#download}{https://hatefulmemeschallenge.com/\#download}}. HarMeme and Harm-P are distributed for research purposes only, without a license for commercial use. 
MultiOFF is licensed under CC-BY-NC. MAMI is under Apache License 2.0. There is no specified license for PrideMM.

\section{Experiment Setup and Implementation Details}
\label{appendix:exp_setup}
\paragraph{Environment.} 
\texttt{PyTorch 2.5.1}, \texttt{CUDA 12.4}, Huggingface \texttt{Transformer 4.45.0 } and \texttt{Python 3.10.12} were used for implementing the experiments. FAISS \cite{johnson_Faiss2019billion} vector similarity search library with version \texttt{faiss-gpu 1.7.2} was used to perform dense retrieval.  All the reported metrics were computed by \texttt{TorchMetrics 1.0.1}. 
\paragraph{Implementation Details.} We use QLoRA \cite{Dettmers2023qlora} to fine-tune all LMMs, as our experiments show that LoRA and QLoRA perform similarly on this task while significantly outperforming full-parameter fine-tuning. The details for fine-tuning are covered in Appendix~\ref{appendix:LMM_experiments}. All reported metrics were based on the mean of five runs with different seeds. For statistical significance testing, each model is run five times with different random seeds. For baseline models, we strictly follow the settings specified in their original papers.
\paragraph{Implementation environment.}
We conducted our experiments on a workstation equipped with an NVIDIA RTX 3090.
%The full parameter fine-tuning experiments were carried out on 4 A100-80GB-SXM GPUs. 

\paragraph{Run time}
\label{appendix:run_time}
The run time for RA-HMD two-stage fine-tuning on the HatefulMemes dataset is approximately 4 hours on a single NVIDIA RTX 3090 GPU, and costs around 1 USD.

To optimize efficiency in stage 2, we pre-extract the final hidden states from the frozen LMM and store them on disk before training, avoiding redundant LMM computations. This reduces the stage 2 training time to approximately 10 minutes.

In our ablation study, we examine the performance impact of merging the two-stage loss into a single fine-tuning stage. Since the LMM remains trainable in this setting, we cannot precompute and store the frozen LMM features, leading to significantly higher computational costs. This approach requires approximately 12 hours to complete fine-tuning on a single RTX 3090.

%For full-parameter fine-tuning, training takes 6 hours on 4 A100-80G.

\subsection{LLaVA and Qwen2-VL experiments}
\label{appendix:LMM_experiments}
We freeze the vision module throughout fine-tuning, following the standard LMM fine-tuning protocol. For prompt formatting, we adhere to InstructBLIP~\cite{DaiInstructBLIP2023}. For LLaVA few-shot experiments, since LLaVA is not explicitly trained to support in-context learning, we follow the procedure outlined by \citet{zong2024vlicl} to enable few-shot learning on LLaVA.
For fine-tuning LLaVA~\cite{LiuLLAVA2023, Liu_2023_LLAVA1.5}, we follow the original hyperparameters setting\footnote{\href{https://github.com/haotian-liu/LLaVA}{https://github.com/haotian-liu/LLaVA}} for fine-tuning on downstream tasks for both the SFT and RA-HMD stage 1 fine-tuning.

For Qwen2-VL fine-tuning, we employ the officially recommended fine-tuning library \texttt{LLaMA-Factory 0.9.1}\footnote{\href{https://github.com/hiyouga/LLaMA-Factory}{https://github.com/hiyouga/LLaMA-Factory}} with official hyperparameter settings for downstream tasks in both the SFT and RA-HMD stage 1 fine-tuning. The only modifications are the LoRA hyperparameters: we employ a larger rank (64) and alpha (128), which are kept fixed throughout all experiments. We fix these LoRA hyperparameters throughout. 

Note that in the open-source version on GitHub \footnote{\href{https://github.com/JingbiaoMei/RGCL}{https://github.com/JingbiaoMei/RGCL}}, we update the LLaMA-Factory version to support Qwen2.5-VL fine-tuning for later study, but the reported results in this paper are based on the version noted above. We further provide scripts to run Qwen2.5-VL-7B and Qwen2.5-VL-32B experiments. Surprisingly, both these two models achieve almost identical results compared to the Qwen2-VL-7B model reported in this paper. Given that, we do not further scale up the parameter to 72B for this study. 

For few-shot learning with Qwen2-VL, we follow the official multi-round conversation prompt format to ensure consistency with the model’s intended usage.

\subsection{Hyperparameters for MLP and Stage 2 Fine-tuning}
\label{appendix:hyperparam}
The default hyperparameters for the MLP and the stage 2 contrastive fine-tuning are shown in Table~\ref{tab:hyperparameters}. With this configuration of hyperparameters, the number of trainable parameters is about 5 million. 
\begin{table}[h]
\small
\centering
\begin{tabular}{ll}
\toprule
Modelling Hyperparameter & Value \\
\midrule
Projection dimension of MLP & 1024 \\
Number of layers in the MLP & 2 \\
Optimizer & AdamW \\
Maximum epochs & 30 \\
Batch size & 64 \\
Learning rate & 0.0001 \\
Weight decay & 0.0001 \\
Gradient clip value & 0.1 \\
\midrule 
Stage 2 Contrastive Learning Hyperparameter  & Value \\
\midrule
\# hard negative examples & 1 \\
\# pseudo-gold positive examples & 1 \\
Similarity metric & Cosine similarity \\
Loss function & NLL \\
Top-K for RKC & 20\\ 

\bottomrule
\end{tabular}
\caption{Default hyperparameter values}
\label{tab:hyperparameters}
\end{table}

\section{Statistical Significance Test}
\label{appendix:statistical_sig}
We conduct statistical significance tests comparing the performance of SFT and RA-HMD fine-tuned LMMs, as reported in Table~\ref{tab:results_p}.

\begin{table*}[h]
\sisetup{table-format=2.1, table-auto-round=true, table-number-alignment=center, detect-weight=true, detect-family=true,mode=text}
\small
\centering
\setlength{\tabcolsep}{4pt}{
\resizebox{1.0\linewidth}{!}{%
\begin{tabular}{ll|SS|SS|SS|SS|SS|SS}
\toprule 
\multicolumn{2}{c|}{} &
 \multicolumn{2}{c|}{\textbf{HatefulMemes}} & \multicolumn{2}{c|}{\textbf{HarMeme}} & \multicolumn{2}{c|}{\textbf{MAMI}} & \multicolumn{2}{c|}{\textbf{Harm-P}} & \multicolumn{2}{c|}{\textbf{MultiOFF}} & \multicolumn{2}{c}{\textbf{PrideMM}}  \\
 & Model  & \textbf{AUC} &  \textbf{Acc.} & \textbf{AUC} & \textbf{Acc.} & \textbf{AUC} &  \textbf{Acc.} & \textbf{Acc.} & \textbf{F1} & \textbf{Acc.} & \textbf{F1} & \textbf{Acc.} & \textbf{F1} \\ 
\midrule
\multicolumn{2}{c|}{LLaVA-1.5-7B} & \multicolumn{2}{c|}{} & \multicolumn{2}{c|}{} & \multicolumn{2}{c|}{}& \multicolumn{2}{c|}{}& \multicolumn{2}{c|}{} \\
& ~\textit{p}-value & \multicolumn{1}{c}{9.8$e^{-3}$} &  \multicolumn{1}{c|}{3.5$e^{-3}$} & \multicolumn{1}{c}{1.2$e^{-2}$} & \multicolumn{1}{c|}{8.5$e^{-3}$} & \multicolumn{1}{c}{4.4$e^{-3}$} & \multicolumn{1}{c|}{6.2$e^{-3}$} & \multicolumn{1}{c}{2.5$e^{-3}$} & \multicolumn{1}{c|}{1.6$e^{-3}$} & \multicolumn{1}{c}{6.1$e^{-3}$} & \multicolumn{1}{c|}{4.6$e^{-3}$} & \multicolumn{1}{c}{5.6$e^{-3}$} & \multicolumn{1}{c}{8.9$e^{-3}$} \\ 
 \multicolumn{2}{c|}{Qwen2-VL-2B} & \multicolumn{2}{c|}{} & \multicolumn{2}{c|}{} & \multicolumn{2}{c|}{}& \multicolumn{2}{c|}{}& \multicolumn{2}{c|}{} \\
& ~\textit{p}-value & \multicolumn{1}{c}{4.2$e^{-3}$} &  \multicolumn{1}{c|}{4.8$e^{-3}$} & \multicolumn{1}{c}{9.1$e^{-3}$} & \multicolumn{1}{c|}{7.6$e^{-3}$} & \multicolumn{1}{c}{6.5$e^{-4}$} & \multicolumn{1}{c|}{1.9$e^{-4}$} & \multicolumn{1}{c}{9.3$e^{-4}$} & \multicolumn{1}{c|}{1.1$e^{-3}$} & \multicolumn{1}{c}{2.0$e^{-2}$} & \multicolumn{1}{c|}{7.7$e^{-3}$} & \multicolumn{1}{c}{7.1$e^{-3}$} & \multicolumn{1}{c}{8.3$e^{-3}$} \\ 
 \multicolumn{2}{c|}{Qwen2-VL-7B} & \multicolumn{2}{c|}{} & \multicolumn{2}{c|}{} & \multicolumn{2}{c|}{}& \multicolumn{2}{c|}{}& \multicolumn{2}{c|}{} \\

%& Average gain $\uparrow$& +4.56  & +2.87 & +2.07 & +5.49 & +8.17 & +8.04  &  +5.74 & +5.46 & +2.81 & +6.29 & +2.05 & +1.60\\  
& ~\textit{p}-value & \multicolumn{1}{c}{8.7$e^{-4}$} & \multicolumn{1}{c|}{2.4$e^{-3}$} & \multicolumn{1}{c}{3.5$e^{-2}$} & \multicolumn{1}{c|}{1.6$e^{-2}$} & \multicolumn{1}{c}{2.5$e^{-3}$} & \multicolumn{1}{c|}{2.6$e^{-3}$} & \multicolumn{1}{c}{6.3$e^{-3}$} & \multicolumn{1}{c|}{8.6$e^{-3}$} & \multicolumn{1}{c}{1.3$e^{-2}$} & \multicolumn{1}{c|}{5.9$e^{-3}$} &\multicolumn{1}{c}{9.2$e^{-3}$} & \multicolumn{1}{c}{7.2$e^{-3}$}\\

\bottomrule
\end{tabular}
}}
\caption{
For each LMM, we provide the ~\textit{p}-value from significance testing between SFT and RA-HMD. }
\label{tab:results_p}
\end{table*}

\section{Insights for using LMMs representation for Meme Classification}
\label{appendix:insights_representation}
There has been substantial interest in adapting decoder-only language models for retrieval and classification tasks. In this section, we summarize the novelty of our approach in comparison to previous efforts that attempt to repurpose decoder-only large language models (LLMs) or large multimodal models (LMMs) for such tasks.

We categorize our adaptation attempts into two groups: (1) those that entirely failed to work for hateful meme classification, and (2) those that showed some promise in improving classification performance but significantly compromised the model’s language generation capabilities. Finally, we explain the rationale that led to the design of our current RA-HMD architecture.
\\
\textbf{Attempts That Do Not Work}
\\
Some prior approaches aim to modify the model architecture with contrastive training objective to enable bidirectional representations from decoder-only models. 

For instance, LLM2Vec \cite{llm2vec2024} proposes to introduce bidirectional attention and masked next-token prediction to make decoder-only models suitable for retrieval and general-purpose text embeddings. 

AutoVER \cite{Xiao2024AutoVER}, on the other hand, introduces a special [RET] token and learns its embedding for visual entity retrieval. 

However, neither approach yielded meaningful improvements in our setting. When applied to hateful meme classification, both methods achieved classification accuracies only slightly above 60\%. We attribute this failure to the limited availability of hateful meme training data, which is insufficient for training these complex adaptations effectively.
\\
\textbf{Approaches That Partially Work}
\\
Other works have explored using decoder-only LLM or LMM embeddings by either pooling the output token representations or using the final-layer embedding of the last token \cite{zhang2024gmeQwen, li2023GTE, Wang2024ImproveTextEmbed}, often in combination with contrastive learning. 

In our experiments, we found that mean pooling generally underperformed compared to using the final token embedding. While this approach achieved a reasonable classification accuracy of about 77\% on the HatefulMemes dataset using Qwen2-VL-7B (RA-HMS Qwen2-VL-7B has an accuracy of 82\%), it had a major drawback: the language generation ability of the model was completely compromised. In practice, the model became unable to generate coherent text, suggesting that the learned representations were no longer aligned with the original language modeling objective.
\\
\textbf{Our approach}
\\
To address these limitations, we propose two key improvements that underpin the RA-HMD architecture:
\begin{itemize}
    \item \textbf{MLP Projection Head}: We introduce a lightweight, trainable multilayer perceptron (MLP) on top of the final token embedding from the LMM. While the language model head (LMH) continues to use the original last-token embedding for language generation, the classification and retrieval heads operate on the MLP-projected embedding. This separation enables the model to retain its language generation capability while learning representations that are better suited for classification and retrieval.
    \item \textbf{Preserving the Language Modeling Objective}: During Stage 1 of RA-HMD fine-tuning, we retain the original language modeling loss in the training objective. This encourages the base embeddings to remain useful for text generation, avoiding overfitting solely to the classification task and preserving the model’s general-purpose functionality.
\end{itemize}

\section{Numbers of Shots and Neighbors}
\label{appendix:ablate_icl_rkc}
We ablate the effects of varying the number of shots for few-shot in-context learning and varying the number of top K nearest neighbors for RKC.

\begin{figure}[h]
    \centering    \includegraphics[width=\linewidth]{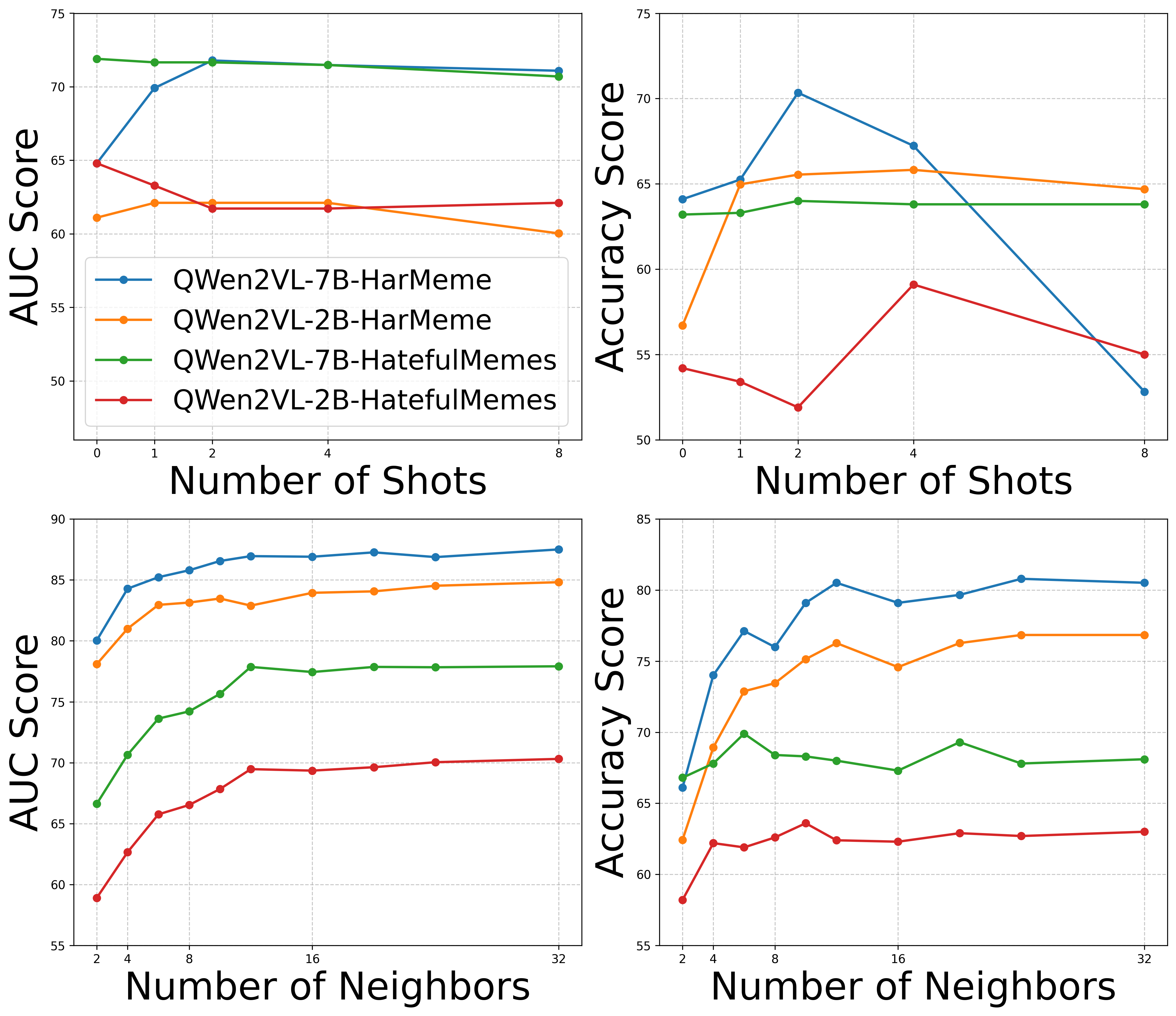}
    \caption{Effects of increasing number of shots for in-context learning with pre-trained LMM and effects of increasing top K nearest neighbors for RKC trained with RA-HMD }
    \label{fig:abl_shots_topk}
\end{figure}

Figure~\ref{fig:abl_shots_topk} demonstrates that increasing the number of in-context examples for LMMs does not consistently yield performance improvements over the zero-shot setting, and in some cases even causes loss.
These findings suggest that merely adding more shots does not necessarily improve performance, which is consistent with findings from \citet{Huang_LowResourceLMMAgentHatefulMeme_2024}. 

Figure~\ref{fig:abl_shots_topk} shows that as the number of nearest neighbors $K$ for RKC increases, the performance continues to increase for both AUC and accuracy, plateauing at around $K=20$. The consistent improvement in performance indicates that RKC trained with RA-HMD utilizes demonstration examples more effectively than the standard in-context learning framework.

\section{Comparing Out-of-Domain Generalization Across Model Variants with In-Context Learning and RKC} 
\label{appendix:rkc_icl_ft}
In this section, we compare the performance of the RKC inference mode against few-shot in-context learning for pre-trained LMMs, SFT LMMs, and LMMs fine-tuned using our proposed RA-HMD framework under the cross-dataset setting in Table~\ref{tab:ablation_rkc_lmms}. 
We observe three things:
\begin{itemize}
    \item RA-HMD-trained Qwen2-VL-7B exhibits more robust generalization in the cross-dataset setting. Its in-context learning performance in few-shot scenarios consistently surpasses that of other models.
    \item RKC consistently outperforms the in-context learning approach across all LMM variants, demonstrating its superior ability to leverage demonstration examples. 
    \item Moreover, RA-HMD fine-tuned LMMs with RKC outperform SFT LMMs with RKC, highlighting the effectiveness of our fine-tuning strategy.
\end{itemize}

\begin{table}[h]
\centering
\small
\resizebox{1.0\linewidth}{!}{%
\begin{tabular}{ll|ll|ll}
\toprule
 & &\multicolumn{2}{c|}{\textbf{HatefulMemes}} & \multicolumn{2}{c}{\textbf{HarMeme}}\\
 Model & Mode & \textbf{AUC} & \textbf{Acc.} & \textbf{AUC} & \textbf{Acc.} \\ \midrule
Pre-trained & Few-shot & 71.5 & 63.8 & 71.5 & 67.2 \\
Pre-trained & RKC & 74.5 & 64.5 & 80.1 & 72.4 \\
SFT & Few-shot & 72.3 & 60.6 & 67.2 & 62.4 \\
SFT & RKC & 75.8 & 67.1 & 84.5 & 75.4 \\
RA-HMD&Few-shot & 74.3 & 63.5 & 73.2 & 68.1 \\
RA-HMD&RKC & 77.1 & 69.3 & 88.8 & 81.7\\ 

\bottomrule
\end{tabular}
}
\caption{Comparing Pre-trained, SFT and RA-HMD systems with few-shot learning and RKC with Qwen2-VL-7B under cross-dataset settings. See Table~\ref{tab:results_low_resource}}
\label{tab:ablation_rkc_lmms}
\end{table}

\section{Comparing Different Inference Modes}
\label{appendix:abl_infer_mode}
%In our experiments the default inference mode varies by setting: for the pre-trained and SFT models, we use the \howard{LM Head} (LMH), whereas for RA-HMD under supervised settings, we use logistic regression (LR). The Retrieval-augmented KNN Classifier (RKC) is employed as the default for RA-HMD in cross-dataset settings. 
Table~\ref{tab:ablation_modes} compares Qwen2-VL-7B fine-tuned with RA-HMD using the three classifiers. Our results indicate that under supervised settings, the differences among the three inference modes are minimal. However, under cross-dataset settings, there is a significant disparity in generalization performance. Notably, RKC outperforms both LMH and LRC, underscoring its superior effectiveness in handling out-of-domain examples.
\begin{table}[htb]
\centering
\begin{subtable}[t]{0.45\textwidth}
\small
\centering
\begin{tabularx}{\textwidth}{l|ll|ll}
\toprule
 &\multicolumn{2}{c|}{\textbf{HatefulMemes}} & \multicolumn{2}{c}{\textbf{HarMeme}}\\
 Inference Mode & \textbf{AUC} & \textbf{Acc.} & \textbf{AUC} & \textbf{Acc.} \\ \midrule
LMH & 90.2 & 81.9 & 92.8 & 88.0 \\
LRC & 91.1 & 82.1 & 93.2 & 88.1 \\
RKC & 90.8 & 81.8 & 93.2 & 88.0 \\
\bottomrule
\end{tabularx}
\caption{Supervised settings, see Table~\ref{tab:results_supervised} for detailed settings}
\label{tab:ablation_mode_supervised}
\vspace{3pt}
\end{subtable}
\hfill
\begin{subtable}[t]{0.45\textwidth}
\small
\centering
\begin{tabularx}{\textwidth}{l|ll|ll}
\toprule
 &\multicolumn{2}{c|}{\textbf{HatefulMemes}} & \multicolumn{2}{c}{\textbf{HarMeme}}\\
 Inference Mode & \textbf{AUC} & \textbf{Acc.} & \textbf{AUC} & \textbf{Acc.} \\ \midrule
LMH & 74.2 & 64.3 & 64.5 & 60.3 \\
LRC & 59.5 & 55.4 & 57.9 & 52.2 \\
RKC & 77.1 & 69.3 & 88.8 & 81.7\\ 
\bottomrule
\end{tabularx}
\caption{Cross-dataset settings, see Table~\ref{tab:results_low_resource} for detailed settings}
\label{tab:ablation_mode_cross}
\end{subtable}
\caption{Comparing different inference modes using RA-HMD fine-tuned Qwen2-VL-7B. RKC shows much better out-of-domain generalization compared to other inference modes.}
\label{tab:ablation_modes}
\end{table}

\section{Ablation study on the loss function}
\label{appendix:abl_loss}
Table~\ref{tab:ablation_loss} shows the results when each loss objective is removed from different stages of fine-tuning. Notably, when the cross-entropy loss is removed in stage 1 for the logistic regression component, the LRC fails to train properly via backpropagation, resulting in performance that is equivalent to random guessing. Consequently, we exclude this case from our comparison. Overall, we observe that removing any loss function from the fine-tuning objective leads to a significant drop in performance, highlighting the importance of each loss term in optimizing the model.

Furthermore, \citet{RGCL2024Mei} utilize in-batch negative examples alongside hard negative examples during training. However, we find that incorporating in-batch negatives in Stage 2 of RA-HMD's contrastive fine-tuning introduces noise and leads to a slight degradation in performance.
\begin{table}[h]
\centering
\begin{subtable}[t]{0.5\textwidth}
\small
\centering
\resizebox{1.0\linewidth}{!}{%
\begin{tabular}{l|ll|ll}
\toprule
 &\multicolumn{2}{c|}{\textbf{HatefulMemes}} & \multicolumn{2}{c}{\textbf{HarMeme}}\\
 Mode & \textbf{AUC} & \textbf{Acc.} & \textbf{AUC} & \textbf{Acc.} \\ \midrule
RA-HMD & \textbf{91.1} & \textbf{82.1} & \textbf{93.2} & \textbf{88.1} \\
~\textit{w/o  $\mathcal{L}^{LM}$ in stage 1} & 88.4 & 79.6 & 90.9 & 85.1 \\
~\textit{w/o  $\mathcal{L}^{CL}$ in stage 2} & 90.2 & 81.2 & 91.9 & 86.4 \\
~\textit{w/o  $\mathcal{L}^{LR}$ in stage 2} & 89.2 & 80.6 & 91.6 & 87.2 \\
\bottomrule
\end{tabular}
}
\caption{Supervised settings, see Table~\ref{tab:results_supervised}}
\label{tab:ablation_loss_supervised}
\vspace{3pt}
\end{subtable}
\hfill
\begin{subtable}[t]{0.5\textwidth}
\small
\centering
\resizebox{1.0\linewidth}{!}{%
\begin{tabular}{l|ll|ll}
\toprule
 &\multicolumn{2}{c|}{\textbf{HatefulMemes}} & \multicolumn{2}{c}{\textbf{HarMeme}}\\
 Mode & \textbf{AUC} & \textbf{Acc.} & \textbf{AUC} & \textbf{Acc.} \\ \midrule
RA-HMD & \textbf{77.1} & \textbf{69.3} & \textbf{88.8} & \textbf{81.7} \\ 
~\textit{w/o  $\mathcal{L}^{LM}$ in stage 1} & 75.4 & 66.6 & 87.3 & 81.1 \\
~\textit{w/o  $\mathcal{L}^{CL}$ in stage 2} & 73.8 & 64.3 & 82.9 & 76.5 \\
~\textit{w/o  $\mathcal{L}^{LR}$ in stage 2} & 76.4 & 67.9 & 86.9 & 80.6 \\
\bottomrule
\end{tabular}
}
\caption{Cross-dataset settings, see Table~\ref{tab:results_low_resource}}
\label{tab:ablation_loss_cross}
\end{subtable}
\caption{Ablation study of RA-HMD two-stage fine-tuning framework on Qwen2-VL-7B, evaluating the impact of removing any of the loss objectives.}
\label{tab:ablation_loss}
\end{table}

\section{Evaluation on General Vision-Language Benchmarks}
\label{appendix:general_VL}
To evaluate general vision-language capabilities, we use models fine-tuned on the HatefulMemes dataset via both supervised fine-tuning (SFT) and our proposed RA-HMD approach.  We conduct the evaluation using the  \texttt{VLMEvalKit} package\footnote{\href{https://github.com/open-compass/VLMEvalKit}{https://github.com/open-compass/VLMEvalKit}} \cite{duan2024vlmevalkit}. For MMMU, we report accuracy based on exact match using the \texttt{MMMU\_DEV\_VAL} split to ensure reproducibility. For SeedBench, we use the \texttt{SeedBench\_IMG} subset, also evaluated with exact match. For GQA, we report results on the \texttt{GQA\_TestDev\_Balanced} split. The full results are provided in Table~\ref{tab:generalVisual_all}.

\begin{table}[h]
\small
\centering
\begin{tabular}{llll}
\toprule
Model & MMMU & SEEDBench & GQA \\
\midrule
Qwen2-VL-2B & 40.2 & 72.7 & 60.4 \\
~\textit{+SFT} & 39.1 & 72.1 & 57.0 \\
~\textit{+RA-HMD} & 40.4 & 72.7 & 60.1\\
\midrule
Qwen2-VL-7B & 49.3 & 76.4 & 62.4 \\
~\textit{+SFT} & 48.0 & 75.2 & 61.2 \\
~\textit{+RA-HMD} & 49.2 & 76.4 & 62.3\\
\bottomrule
\end{tabular}
\caption{Comparison of the pretrained, SFT, and RA-HMD Qwen2-VL-2B models on three general vision-language benchmarks. The SFT and RA-HMD models are fine-tuned on the HatefulMemes dataset.}
\label{tab:generalVisual_all}
\end{table}

\section{Task-specific fine-tuning and Multi-task fine-tuning}
Here, we compare RA-HMD–tuned Qwen2-VL-7B models under two settings: (i) task-specific fine-tuning, where separate models are trained for each dataset, and (ii) multi-task fine-tuning, where a single model is jointly trained on all datasets. As shown in Table~\ref{tab:mtl_vs_stl}, multi-task fine-tuning results on Qwen2-VL-7B with RA-HMD suggest that a single system fine-tuned on all available datasets performs comparably to domain-specific systems.
\begin{table*}[t]
\centering
\small
\begin{tabular}{lcccccc}
\toprule
\textbf{Model} & \textbf{HatefulMemes} & \textbf{HarMeme} & \textbf{MAMI} & \textbf{Harm-P} & \textbf{MultiOFF} & \textbf{PrideMM} \\
\midrule
Single-task fine-tuning & 82.1 & 88.1 & 79.9 & 91.6 & 71.1 & 78.1 \\
Multi-task fine-tuning        & 82.0 & 87.8 & 80.4 & 91.6 & 70.4 & 78.5 \\
\bottomrule
\end{tabular}

\caption{Accuracy comparison between single-task and multi-task fine-tuning across six datasets.}
\label{tab:mtl_vs_stl}

\end{table*}

\section{GPT-4o Results and Prompts}
\label{appendix:gpt4o}
We adopt the \texttt{gpt-4o-2024-05-13} model for our evaluations. We explore two prompting strategies for GPT-4o. In Table~\ref{tab:results_supervised}, we present results using optimized prompts, where the specific type of hate (e.g., hatefulness, harmfulness, offensiveness, misogyny, etc.) is explicitly provided based on the characteristics of each dataset. Below, we include the detailed prompt formulations used for each dataset:
\begin{itemize}
    \item \textbf{HatefulMemes}:\\
    \textit{You are an expert in identifying meme-based social abuse. Given the meme, with the Text [T] accompanied by the image. The task is to classify whether the meme contains hateful content or not. Please answer with 'yes' or 'no'.}
    
    \item \textbf{Harmeme}:\\
    \textit{You are an expert in identifying meme-based social abuse. Given the meme, with the Text [T] accompanied by the image. The task is to classify whether the meme contains harmful content or not. Please answer with 'yes' or 'no'.}
    
    \item \textbf{Harmp}:\\
    \textit{You are an expert in identifying meme-based social abuse. Given the meme, with the Text [T] accompanied by the image. The task is to classify whether the meme contains harmful content or not. Please answer with 'yes' or 'no'.}
    
    \item \textbf{MultiOFF}:\\
    \textit{You are an expert in identifying meme-based social abuse. Given the meme, with the Text [T] accompanied by the image. The task is to classify whether the meme contains offensive content or not. Please answer with 'yes' or 'no'.}
    
    \item \textbf{PrideMM}:\\
    \textit{You are an expert in identifying meme-based social abuse. Given the meme, with the Text [T] accompanied by the image. The task is to classify whether the meme contains hateful content related to LGBTQ+ Pride movement or not. Please answer with 'yes' or 'no'.}
    
    \item \textbf{MAMI}:\\
    \textit{You are an expert in identifying meme-based social abuse. Given the meme, with the Text [T] accompanied by the image. The task is to classify whether the meme contains misogyny or not. Please answer with 'yes' or 'no'.}
\end{itemize}

The low-resource comparison in Table~\ref{tab:results_low_resource} is designed to reflect real-world scenarios to detect the evolving harmful memes on the internet, where the specific type of hate is often unknown. Accordingly, we use the general term “hate” across all six datasets in this setting. Below, we include the detailed prompt formulations:
\begin{itemize}
    \item Given the meme, with the Text [T] accompanied by the image. Does the meme contain any hateful content or any social abuse?
\end{itemize}
We directly present the comparison between the two sets of results in Table~\ref{tab:results_compare_gpt}. For reference, Goat-Bench \cite{Lin2025GoatBench} published GPT-4o results on similar datasets using task-specific prompts. However, their evaluation is based on different data splits, making the results not directly comparable. Below, we summarize key differences in performance:

\begin{itemize}
\item \textbf{Hatefulness}: This benchmark corresponds to our HatefulMemes dataset. While we use the \texttt{test\_seen} split, Goat-Bench uses the \texttt{test\_unseen} split. Despite this difference, the results are comparable: they report an accuracy of 71.7, while ours is 71.3.
\item \textbf{Harmfulness}: This benchmark aligns with our combined evaluation on HarMeme and Harm-P. Goat-Bench reports an accuracy of 66.01, whereas we achieve 72.9 on HarMeme and 63.1 on Harm-P.

\item \textbf{Offensiveness}: Due to the small size of the MultiOFF dataset, Goat-Bench evaluates performance using the combined training, validation, and test sets. In contrast, we report results based solely on the test set. They report an accuracy of 62.13 and an F1 score of 61.16, compared to our accuracy of 58.3 and F1 score of 58.1.
\end{itemize}

\begin{table*}[h]
\small
\sisetup{table-format=2.1, table-auto-round=true, table-number-alignment=center, detect-weight=true, detect-family=true,mode=text}
\centering
\setlength{\tabcolsep}{5pt}{
\resizebox{1.0\linewidth}{!}{%
\begin{tabular}{ll|SS| SS| SS| SS| SS| SS}
\toprule 
&&
 \multicolumn{2}{c|}{\textbf{HatefulMemes}} & \multicolumn{2}{c|}{\textbf{HarMeme}} & \multicolumn{2}{c|}{\textbf{MAMI}} & \multicolumn{2}{c|}{\textbf{Harm-P}} & \multicolumn{2}{c|}{\textbf{MultiOFF}} & \multicolumn{2}{c}{\textbf{PrideMM}}  \\
 &Model  & \textbf{AUC} &  \textbf{Acc.} & \textbf{AUC} & \textbf{Acc.} & \textbf{AUC} &  \textbf{Acc.} & \textbf{Acc.} & \textbf{F1} & \textbf{Acc.} & \textbf{F1} & \textbf{Acc.} & \textbf{F1} \\ 
\midrule
& GPT-4o (Specific Prompt) & \text{-} & 71.3 & \text{-} & 72.9 & \text{-} & 79.4 & 63.1 & 64.5 & 58.3 & 58.1 & 75.3 & 73.7  \\
 & GPT-4o (General Prompt) & \text{-} & 66.40 & \text{-}  & 68.4 & \text{-} & 72.90 & 55.39 & 55.11 & 61.07 & 51.07 & 63.79 & 62.28   \\ % add to both tables 

\bottomrule
\end{tabular}
}}
\caption{Comparing the performance with GPT-4o with different prompts }
\label{tab:results_compare_gpt}
\end{table*}

\section{Rationale Generated by LMMs}
\label{appendix:rationale}
We compare meme explanations generated by the SFT and RA-HMD fine-tuned Qwen2-VL-7B models (Table~\ref{tab:results_supervised} rows 15 and 16). RA-HMD is evaluated after Stage-1 fine-tuning, as Stage-2 does not fine-tune the Language Model Head for language generation.
For evaluation, we prompt the above two models with:

\textit{ "Does the meme contain any hate speech or offensive content? Please provide a detailed explanation."}

Following prior work \cite{Yang2023Hare}, we assess explanation quality using an LLM judge. Specifically, we provide GPT-4o-mini (\texttt{gpt-4o-mini-2024-07-18}) with reference explanations from \cite{Hee_DecodeMeaningFBHM_Hatred_2023} and use the comparison template from \cite{Yang2023Hare} for pairwise evaluation. For reference, the prompt we use is:

\begin{lstlisting}[label={lst:prompt_judge}]
Please act as an impartial judge and evaluate the quality of the model-generated reasoning provided by two AI assistants. You will compare the two model-generated reasoning with the reference human reasoning for a hateful meme.

Your evaluation should consider which response is more similar to the true answers. Begin your evaluation by comparing the two responses and provide a short explanation. After providing your explanation, output your final verdict by strictly following this format: "[[A]]" if assistant A is more accurate, "[[B]]" if assistant B is more accurate, and "[[C]]" for a tie.


[The Start of Assistant A's Answer]
{assistant_a["reasoning"]}
[The End of Assistant A's Answer]

[The Start of Assistant B's Answer]
{assistant_b["reasoning"]}
[The End of Assistant B's Answer]}

[The Start of Reference Answer]
{reference}
[The End of Reference Answer]


\end{lstlisting}

Results are as follows:
\begin{itemize}
    \item RA-HMD beats SFT: 61.5% 
    \item RA-HMD ties SFT: 13.8%
    \item SFT beats RA-HMD: 24.7%
\end{itemize}

Based on the analysis of the generated explanations, we find that improvements in classification accuracy are supported by deeper semantic understanding of memes. For challenging examples, where comprehension of background events or fine-grained details from the image is required, the RA-HMD fine-tuned system generates better explanations. 

Additionally, we conduct a rubric-based analysis, following the procedure in \citealt{Yang2023Hare}. Specifically, we prompted the LLM judge to assign a score from 0 to 10 to each system’s rationale, based on how well it aligns with the human-written explanations.

For reference, the prompt we use is:

\begin{lstlisting}[label={lst:prompt_judge}]
Compare the model-generated reasoning with the reference human reasoning for this hateful meme.

Reference: {reference_reasoning}
Model: {model_reasoning}

Rate how well the model reasoning aligns with the reference on a scale of 0-10:
- 9-10: Excellent alignment, captures all key points
- 7-8: Good alignment, captures most key points  
- 5-6: Satisfactory alignment, captures some key points
- 3-4: Poor alignment, misses many key points
- 1-2: Very poor alignment, minimal understanding
- 0: Completely wrong or unrelated

Score: [0-10]
Explanation: [1-2 sentences]
\end{lstlisting}

We assess the SFT and RA-HMD tuned Qwen2-VL-7B on the dev-seen split of the Facebook HatefulMemes dataset. Our results show that the SFT baseline received an average score of 4.9, while RA-HMD scored higher at 5.6, further indicating stronger alignment with human rationales.

For reference, we include sample explanations of the meme examples in the Case Analysis in Appendix~\ref{appendix:case_analysis} Table~\ref{tab:case}.
\\
\textbf{Case 1 (Meme references 9/11 in a political context)}
\begin{itemize}
    \item SFT: No. It does not contain any hate speech or offensive content. It appears to be a political statement questioning the decision to vote for a particular political figure or party. It suggests a critique of the decision to support a political figure or party that is associated with the events mentioned in the text.
    \item RA-HMD: Yes. The meme contains hate speech. It is offensive and disrespectful to suggest that the Twin Towers should be voted into the government. It is a tragic event that should never be forgotten, and it is not appropriate to use it as a political meme.
\end{itemize}
\textbf{Case 2 (Meme associates modern Japanese people with suicide) }
\begin{itemize}
    \item SFT : No. It does not contain any hate speech or offensive content. It simply features a person holding a gun to their head with the text "modern japanese" and two quotation marks. The text and imagery do not convey any offensive or hateful messages.
    \item RA-HMD: Yes. The meme contains hate speech. It is a hateful meme that associates modern Japanese people with suicide. It is a very offensive meme.
\end{itemize}

\textbf{Case 3 (Meme objectifies a woman with references to the disability)}
\begin{itemize}
    \item SFT: No. It does not contain any hate speech or offensive content. The meme contains text that implies a sexual service is available for purchase, which could be interpreted as objectifying or degrading. However, the image itself does not contain any explicit or offensive content.
    \item RA-HMD: Yes. The meme contains hate speech. It is a hateful meme that makes fun of a woman's disability. It is a very offensive meme.

\end{itemize}
These results further demonstrate the superior meme understanding capabilities of RA-HMD, especially when compared to SFT.

\begin{table*}[hb]
    \small
    \centering

    \begin{tabularx}{\textwidth}{Xccc}
    \toprule
        & Case 1 & Case 2 & Case 3 \\
        \midrule
        Meme &\includegraphics[valign=c, width=0.265\textwidth]{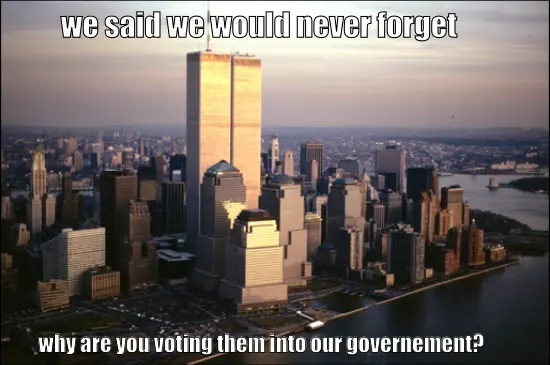} & 
        \includegraphics[valign=c, width=0.26\textwidth]{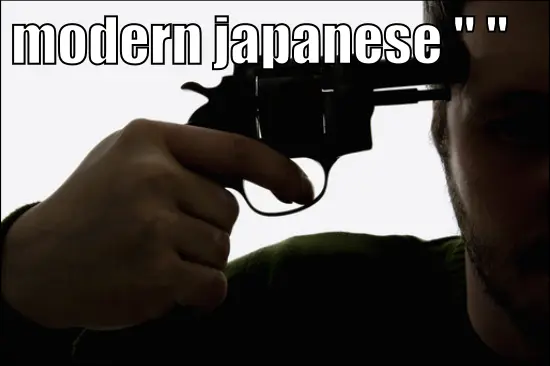} &  
        \includegraphics[valign=c, width=0.22\textwidth]{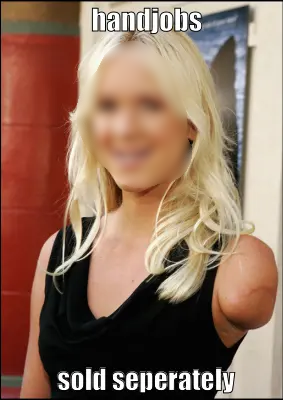}   \\
        \midrule
        Ground Truth &  \#Hateful & \#Hateful & \#Hateful  \\
        SFT  &  \#Benign & \#Benign & \#Benign \\
        RA-HMD  &  \#Hateful & \#Hateful & \#Hateful\\
        \toprule
        &Case 4 & Case 5 & Case 6 \\
        \midrule
        Meme &\includegraphics[valign=c, width=0.26\textwidth]{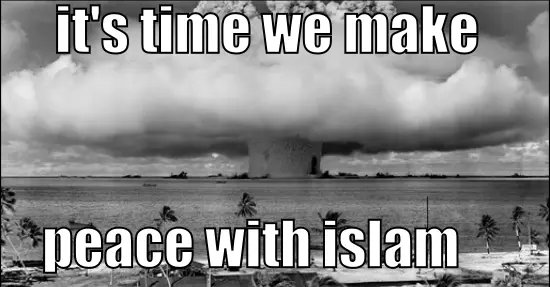} &  
        \includegraphics[valign=c, width=0.26\textwidth]{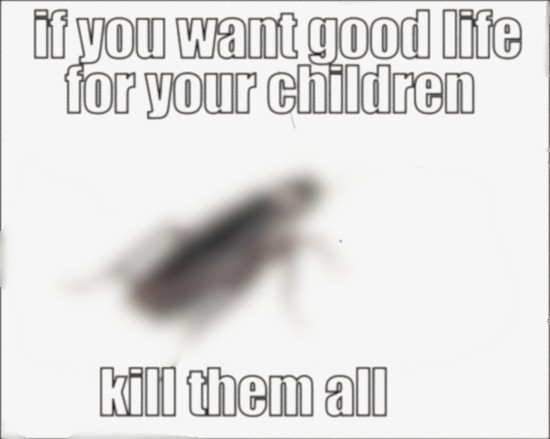} &  
        \includegraphics[valign=c, width=0.22\textwidth]{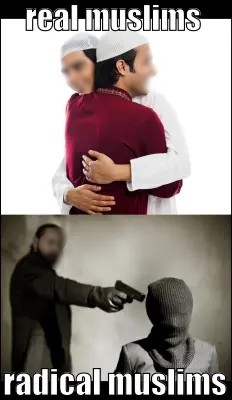}   \\
        \midrule
        Ground Truth &  \#Hateful &  \#Benign & \#Benign  \\
        SFT  &  \#Benign & \#Hateful & \#Hateful \\
        RA-HMD  &  \#Hateful & \#Benign & \#Benign \\
        \bottomrule
    \end{tabularx}
    \caption{Visualization of cases from SFT Qwen2-VL-7B and RA-HMD Qwen2-VL-7B Models on the HatefulMemes Dataset. Case 5 contains an insect in the meme; we applied a blurring filter to obscure it. Furthermore, faces are also blurred. }
    \label{tab:case}
\end{table*}

\begin{table*}[h]
    \small
    \centering

    \begin{tabularx}{\textwidth}{Xccc}
    \toprule
        & Case 1 & Case 2 & Case 3 \\
        \midrule
        Meme &\includegraphics[valign=c, width=0.26\textwidth]{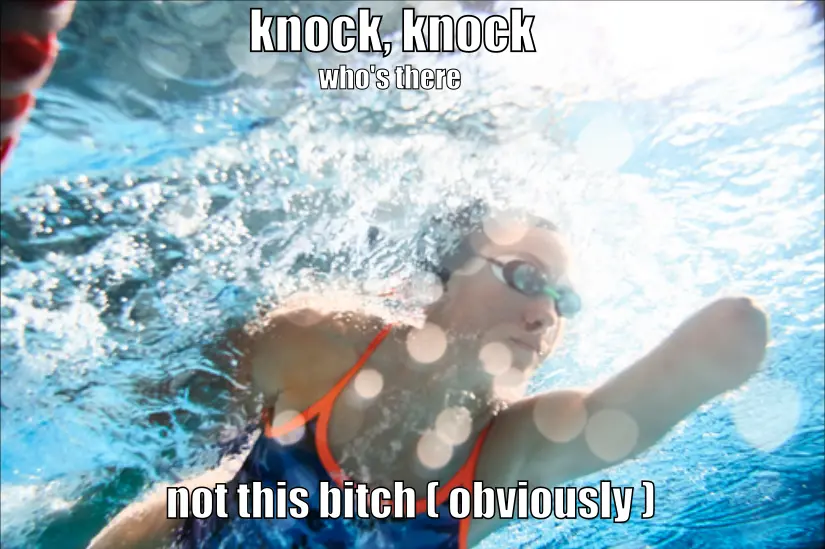} & 
        \includegraphics[valign=c, width=0.26\textwidth]{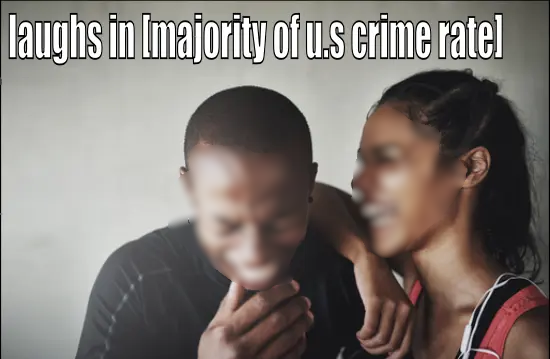} &  
        \includegraphics[valign=c, width=0.26\textwidth]{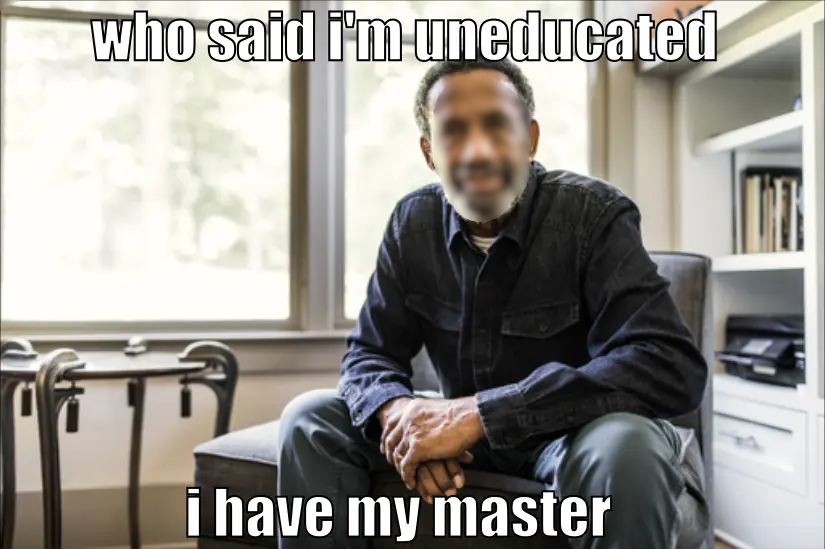}   \\
        \midrule
        Ground Truth &  \#Hateful & \#Hateful & \#Hateful  \\
        SFT  &  \#Benign & \#Benign & \#Benign \\
        RA-HMD  &  \#Benign & \#Benign & \#Benign\\
        \bottomrule
    \end{tabularx}
    \caption{The error cases of SFT Qwen2-VL-7B and RA-HMD Qwen2-VL-7B models on HatefulMemes dataset}
    \label{tab:error_case}
\end{table*}

\section{Case Analysis}
\label{appendix:case_analysis}
\subsection{Comparing SFT and RA-HMD Predictions}
Table~\ref{tab:case} presents examples where our RA-HMD method successfully corrects prediction errors made by the SFT model on Qwen2-VL-7B. Cases 1-4 involve hateful memes, while Cases 5-6 are benign memes that the SFT model misclassified, primarily due to poor multimodal alignment. These examples require a deep, joint understanding of both the image and text, a challenge that our RA-HMD effectively addresses. For example, in Case 2, the model needs to use its understanding of Japanese culture and associate this knowledge with the visual cues in the image.

\subsection{Error Analysis}
\label{appendix:error_analysis}
In Table~\ref{tab:error_case}, we present examples where RA-HMD was unable to correct errors made by the baseline SFT model. In the first case, the model struggles with the nuanced visual understanding required to interpret the disabled body of the swimmer. Additionally, these examples demand complex reasoning to assess the hatefulness of the memes. Interpreting such nuanced meanings remains a challenge for current models. However, we anticipate that the advanced reasoning capabilities of emerging systems like OpenAI-o1 \cite{OpenAI2024o1} and DeepSeek-R1 \cite{DeepSeekAI2025_r1} will help address these limitations.

\section{Baseline Methods}
\label{appendix:baseline_models}
\begin{itemize}
    \item \textbf{Visual Programming Distillation (VPD)} \cite{Hu_2024_VPD} builds an agentic LMM framework by fine-tuning the model's ability to use external tools (e.g., writing and executing programs). VPD fine-tunes PaLI-X 55B, achieving state-of-the-art performance on the HatefulMemes dataset.
    \item \textbf{ISSUES} \cite{Burbi_2023_Issues} employs text inversion along with several projection layers and a feature combiner to enhance the pre-trained CLIP encoder, yielding state-of-the-art results on the HarMeme dataset. 
    \item \textbf{RGCL} \cite{RGCL2024Mei} learns hate-aware vision and language representations through a contrastive learning objective applied to a pre-trained CLIP encoder, achieving state-of-the-art performance on the MultiOFF dataset.
    \item \textbf{ExplainHM} \cite{Lin_2024_ExplainHM} fine-tunes three LLMs arranged as two debaters (arguing whether a meme is hateful) and one judge (summarizing the debaters’ points) to both explain and classify hateful memes.
    \item \textbf{Pro-Cap} \cite{Cao_2023_ProCap} employs prompting techniques to guide pre-trained vision-language models in generating image captions that reflect hateful content. These generated captions are then combined with textual information to improve hateful meme detection.
    \item \textbf{MemeCLIP} \cite{Shah2024memeclip_pridemm} utilizes CLIP features along with feature adapters to mitigate overfitting and employs a cosine classifier to address class imbalance.
    %\item \textbf{MOMENTA} \cite{PramanickMomenta2021} leverages trainable fusion layers—such as Cross-Modal Attention Fusion—to integrate multimodal features extracted by CLIP for improved hateful meme detection.
    \item \textbf{HateCLIPper} \cite{KumarHateClip2022} explores various strategies to align and fuse the visual and textual modalities in CLIP-based encoders, enhancing their performance on challenging hateful meme cases.
    \item \textbf{LOREHM} \cite{Huang_LowResourceLMMAgentHatefulMeme_2024} adopts an agent-based LMM framework that leverages few-shot in-context learning and self-improvement capabilities for low-resource hateful meme detection.
    \item \textbf{Mod-Hate} \cite{Cao_2024_ModHate} trains a suite of LoRA modules and utilizes few-shot demonstration examples to train a module composer, which assigns weights to the LoRA modules for effective low-resource hateful meme detection.
    
\end{itemize}

\section{AI Assistance}
Our coding work was assisted by Github Copilot.
OpenAI ChatGPT was only used in proofreading and spell-checking. We claim that the content
presented in this paper was fully original.
\end{document}